\documentclass[accepted, startpage=3939]{uai2025} %

\usepackage[american]{babel}

\usepackage{natbib} %
\usepackage{mathtools} %
\usepackage{booktabs} %
\usepackage{tikz} %
\usepackage{subcaption}
\usepackage[]{xcolor} %
\usepackage{colortbl}
\usepackage{multirow}

\newcommand{\eg}{\textit{e.g.}\xspace}
\newcommand{\etal}{\textit{et~al.}\xspace}

\hypersetup{colorlinks,allcolors=black}

\usepackage{shortex-light}
\usepackage{pctex}

\usepackage{safecolours}
\colorlet{scBlue}{tol-colour1}
\colorlet{scRed}{tol-colour2}
\colorlet{scGreen}{tol-colour3}
\colorlet{scYellow}{tol-colour4}
\colorlet{scCyan}{tol-colour5}
\colorlet{scPurple}{tol-colour6}

\makeatletter
\let\pgfimagex\pgfimage
\renewcommand{\pgfimage}[2][]{\pgfimagex[#1]{\datapath/#2}}
\makeatother

\usepackage{fontawesome5}
\usepackage{url}
\usepackage{natbib}
\usepackage{pgfplots, tikz}
\usetikzlibrary{arrows.meta, chains, decorations.pathreplacing}
\usetikzlibrary{backgrounds, fit}
\usepgfplotslibrary{patchplots, fillbetween}

\usepackage{xspace}
\newcommand{\ie}{\textit{i.e.}\xspace}

\newlength{\figurewidth}
\newlength{\figureheight}

\newcommand{\ours}{BitVI}

\title{Approximate Bayesian Inference via Bitstring Representations}
\author[1]{\href{mailto:<aleksanteri.sladek@aalto.fi>?Subject=Approximate Bayesian Inference via Bitstring Representations UAI 2025 Paper}{Aleksanteri Sladek}{}}
\author[1]{Martin Trapp}
\author[1]{Arno Solin}
\affil[1]{%
    Department of Computer Science\\
    Aalto University\\
    Espoo, Finland
}

\begin{document}
\maketitle

\begin{abstract}
The machine learning community has recently put effort into quantized or low-precision arithmetics to scale large models. This paper proposes performing probabilistic inference in the quantized, discrete parameter space created by these representations, effectively enabling us to learn a continuous distribution using discrete parameters. We consider both 2D densities and quantized neural networks, where we introduce a tractable learning approach using probabilistic circuits. This method offers a scalable solution to manage complex distributions and provides clear insights into model behavior. We validate our approach with various models, demonstrating inference efficiency without sacrificing accuracy. This work advances scalable, interpretable machine learning by utilizing discrete approximations for probabilistic computations.
\looseness-1

\end{abstract}

\section{Introduction}
\label{sec:intro}

Probabilistic inference is central to modern machine learning, providing a principled framework for reasoning under uncertainty. In Bayesian inference, uncertainty is captured through probability distributions over parameters, with Bayes' theorem offering a systematic way to update beliefs with data. However, exact Bayesian inference is often intractable due to the complexity of the integrals involved. Variational inference (VI) \citep{blei2017variational, jordan1999introduction, wainwright2008graphical} is typically employed as a scalable alternative to Markov chain Monte Carlo (MCMC) methods, enabling inference in high-dimensional models. Despite its success, VI relies on continuous parameterizations and often restrictive Gaussian assumptions, which can introduce representational and computational inefficiencies, particularly in large-scale settings.

\begin{figure}[t!]
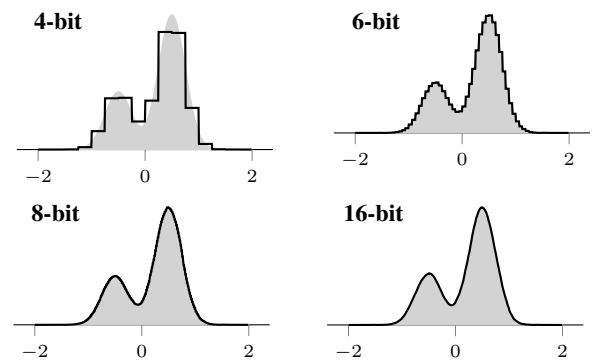

	\centering
	\setlength{\figurewidth}{0.6\columnwidth}
	\setlength{\figureheight}{0.4\columnwidth}
	\pgfplotsset{
      axis y line=none,    %
      axis x line*=bottom,
      tick align=outside,
      ymin=0,
      clip=false,
      ytick=\empty,        %
      yticklabels=\empty,  %
      xticklabel style={font=\scriptsize},
    }
	\newcommand{\mylabel}[1]{%
	  \tikz[baseline]\node[anchor=base,font=\bf\footnotesize,align=left,text width=.6\figurewidth]{#1};\\[-1em]}
	\hfill
    \begin{minipage}[t]{.48\columnwidth}
      \centering\mylabel{4-bit}
      \input{teaser-mixture-4bits} %
    \end{minipage}
    \hfill
    \begin{minipage}[t]{.48\columnwidth}
      \centering\mylabel{6-bit}
	  \input{teaser-mixture-6bits} %
	\end{minipage}
	\hfill\\
	\hfill
    \begin{minipage}[t]{.48\columnwidth}
      \centering\mylabel{8-bit}
      \input{teaser-mixture-8bits} %
    \end{minipage}
    \hfill
    \begin{minipage}[t]{.48\columnwidth}
      \centering\mylabel{16-bit}
      \input{teaser-mixture-16bits} %
    \end{minipage}
	\hfill	
	\caption{\textbf{Capturing a 1D Gaussian mixture} with \ours\ using different numbers of bits in the bitstring. Using 4-bits already yields a reasonable approximation, and the model visually saturates at around 8 bits.}
	\label{fig:teaser}
\end{figure}

\begin{figure*}[t!]
	\centering
	\def\datapath{.}
	\setlength{\figurewidth}{.18\textwidth}
	\setlength{\figureheight}{\figurewidth}
	\pgfplotsset{
	  scale only axis,
	  axis on top,
	  ytick=\empty,        %
	  yticklabels=\empty,  %
	  xtick=\empty,        %
	  xticklabels=\empty,  %
	}

	\begin{subfigure}[t]{.18\textwidth}
	  \centering
\begin{tikzpicture}

\begin{axis}[
height=\figureheight,
tick pos=left,
width=\figurewidth,
xmin=-4, xmax=3.90000000000001,
ymin=-4, ymax=3.90000000000001
]
\addplot graphics [includegraphics cmd=\pgfimage,xmin=-4, xmax=3.90000000000001, ymin=-4, ymax=3.90000000000001] {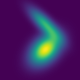};
\end{axis}

\end{tikzpicture}
	\end{subfigure}
	\hfill
	\begin{subfigure}[t]{.18\textwidth}
	  \centering
\begin{tikzpicture}

\begin{axis}[
height=\figureheight,
tick pos=left,
width=\figurewidth,
xmin=-4, xmax=3.90000000000001,
ymin=-4, ymax=3.90000000000001
]
\addplot graphics [includegraphics cmd=\pgfimage,xmin=-4, xmax=3.90000000000001, ymin=-4, ymax=3.90000000000001] {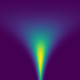};
\end{axis}

\end{tikzpicture}
	\end{subfigure}
	\hfill
	\begin{subfigure}[t]{.18\textwidth}
	  \centering
\begin{tikzpicture}

\begin{axis}[
height=\figureheight,
tick pos=left,
width=\figurewidth,
xmin=-4, xmax=3.90000000000001,
ymin=-4, ymax=3.90000000000001
]
\addplot graphics [includegraphics cmd=\pgfimage,xmin=-4, xmax=3.90000000000001, ymin=-4, ymax=3.90000000000001] {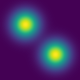};
\end{axis}

\end{tikzpicture}
	\end{subfigure}
	\hfill
	\begin{subfigure}[t]{.18\textwidth}
	  \centering
\begin{tikzpicture}

\begin{axis}[
height=\figureheight,
tick pos=left,
width=\figurewidth,
xmin=-4, xmax=3.90000000000001,
ymin=-4, ymax=3.90000000000001
]
\addplot graphics [includegraphics cmd=\pgfimage,xmin=-4, xmax=3.90000000000001, ymin=-4, ymax=3.90000000000001] {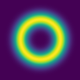};
\end{axis}

\end{tikzpicture}
	\end{subfigure}
	\hfill
	\begin{subfigure}[t]{.18\textwidth}
	  \centering
\begin{tikzpicture}

\begin{axis}[
height=\figureheight,
tick pos=left,
width=\figurewidth,
xmin=-4, xmax=3.90000000000001,
ymin=-4, ymax=3.90000000000001
]
\addplot graphics [includegraphics cmd=\pgfimage,xmin=-4, xmax=3.90000000000001, ymin=-4, ymax=3.90000000000001] {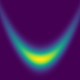};
\end{axis}

\end{tikzpicture}
	\end{subfigure}\\[-.5em]	
		
	\begin{subfigure}[t]{.18\textwidth}
	  \centering
\begin{tikzpicture}

\begin{axis}[
height=\figureheight,
tick pos=left,
width=\figurewidth,
xmin=-4, xmax=3.90000000000001,
ymin=-4, ymax=3.90000000000001
]
\addplot graphics [includegraphics cmd=\pgfimage,xmin=-4, xmax=3.90000000000001, ymin=-4, ymax=3.90000000000001] {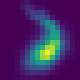};
\end{axis}

\end{tikzpicture}\\[-.5em]
	  \caption{Gaussian Mixture}
	\end{subfigure}
	\hfill
	\begin{subfigure}[t]{.18\textwidth}
	  \centering
\begin{tikzpicture}

\begin{axis}[
height=\figureheight,
tick pos=left,
width=\figurewidth,
xmin=-4, xmax=3.90000000000001,
ymin=-4, ymax=3.90000000000001
]
\addplot graphics [includegraphics cmd=\pgfimage,xmin=-4, xmax=3.90000000000001, ymin=-4, ymax=3.90000000000001] {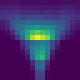};
\end{axis}

\end{tikzpicture}\\[-.5em]
	  \caption{Neal's Funnel}
	\end{subfigure}
	\hfill
	\begin{subfigure}[t]{.18\textwidth}
	  \centering
\begin{tikzpicture}

\begin{axis}[
height=\figureheight,
tick pos=left,
width=\figurewidth,
xmin=-4, xmax=3.90000000000001,
ymin=-4, ymax=3.90000000000001
]
\addplot graphics [includegraphics cmd=\pgfimage,xmin=-4, xmax=3.90000000000001, ymin=-4, ymax=3.90000000000001] {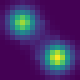};
\end{axis}

\end{tikzpicture}\\[-.5em]
	  \caption{Two-modal Gaussian}
	\end{subfigure}
	\hfill
	\begin{subfigure}[t]{.18\textwidth}
	  \centering
\begin{tikzpicture}

\begin{axis}[
height=\figureheight,
tick pos=left,
width=\figurewidth,
xmin=-4, xmax=3.90000000000001,
ymin=-4, ymax=3.90000000000001
]
\addplot graphics [includegraphics cmd=\pgfimage,xmin=-4, xmax=3.90000000000001, ymin=-4, ymax=3.90000000000001] {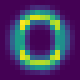};
\end{axis}

\end{tikzpicture}\\[-.5em]
	  \caption{Ring}
	\end{subfigure}
	\hfill
	\begin{subfigure}[t]{.18\textwidth}
	  \centering
\begin{tikzpicture}

\begin{axis}[
height=\figureheight,
tick pos=left,
width=\figurewidth,
xmin=-4, xmax=3.90000000000001,
ymin=-4, ymax=3.90000000000001
]
\addplot graphics [includegraphics cmd=\pgfimage,xmin=-4, xmax=3.90000000000001, ymin=-4, ymax=3.90000000000001] {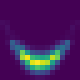};
\end{axis}

\end{tikzpicture}\\[-.5em]
	  \caption{Banana}
	\end{subfigure}
	\caption{Target densities (upper row) and 4-bit \ours\ (lower row) on non-Gaussian 2D density functions. \ours\ can model the densities well despite the low bit precision. We include comparisons to full-covariance Gaussian VI in \cref{fig:densities-app}.}
	\label{fig:2d_densities}
\end{figure*}

To address computational constraints, the machine learning community has increasingly embraced quantization techniques. These methods reduce numerical precision to improve efficiency, leveraging low-bit representations for storage and computation. Many of those can be related to reducing the numerical precision, such as developing tailored low-precision number systems \citep{gustafson2017posit,agrawal2019dlfloat} or methods for parameter quantization. Recent works leveraging large-scale mixed-precision FP8 \citep[\eg,][]{liu2024deepseek}, FP4 \citep{wang2025optimizing}, or even 1-bit neural architectures \citep{ma2024era} have shown innovative low-precision training approaches.

\cref{fig:teaser} illustrates how a Gaussian mixture model, typically represented in high-precision floating point, can be equivalently expressed using a low-precision bitstring representation, motivating the feasibility of inference in quantized spaces. These developments suggest that probabilistic inference need not be tied to continuous-valued computations but can instead be formulated in the space of bitstrings.

This work hinges on the fundamental principle that on a computer, continuous values are represented by finite-length bitstrings---that is, a discrete representation. Hence, probability distributions representable with a computer necessarily possess a representation over finite-length bitstrings. We explore this connection and derive a method to conduct efficient approximate inference through this link.

This work introduces \textbf{BitVI}, a novel approach for approximate probabilistic inference in bitstring models. BitVI exploits the inherent discrete nature of number representations to approximate continuous distributions directly in the space of bitstrings. By leveraging probabilistic circuits \citep{darwiche2003differential,choi2020probabilistic}, our method provides a tractable way to learn and perform inference over complex distributions without requiring high-precision representations. 
\cref{fig:2d_densities} demonstrates how \ours\ can model complex distribution with only 4-bit precision.

We validate BitVI across {\em (i)}~standard benchmark densities, demonstrating its ability to approximate known distributions; 
and {\em (ii)}~Bayesian deep learning in neural network models, where BitVI enables uncertainty quantification. Our results highlight the efficiency and accuracy of BitVI, making it a compelling alternative to traditional inference methods. Our contributions can be summarized as follows.
\begin{itemize}
  \item \textbf{Methodological:} We introduce \ours, a novel approach for approximate Bayesian inference in bitstring models, leveraging probabilistic circuits for efficient learning and inference.
  \item \textbf{Experimental:} We provide proof-of-concept and benchmarking results on standard test problems as well as Bayesian deep learning tasks, demonstrating the effectiveness of \ours\ in practical applications.
  \item \textbf{Insights:} We explore the role of bitstring representations in probabilistic inference and shed light on the trade-offs between model flexibility and quantization.  
\end{itemize}

\begin{figure*}[t!]
	\centering
	\begin{tikzpicture}[nodeDecorate/.style={shape=circle,inner sep=2pt,fill,draw=white,thick}]
  
  		\node[anchor=east] at (-0.25,1.25) {$\hat{q}(b_1, b_2, b_3)=f \Biggl($};
  		\node[anchor=west] at (2.75,1.25) {$\Biggr)$};
  		\node at (1.25,-1) {Distribution over bitstrings $\bb$};

		\foreach \nodename/\x/\y/\label in {
		  1/0/0/000, 2/2/0/001, 6/2.5/0.5/101}
		{
		  \node[label=below:{\scriptsize \label}] (\nodename) at (\x,\y) [nodeDecorate] {};
		}
		\foreach \nodename/\x/\y/\label in {
		 3/2/2/011, 4/0/2/010, 7/2.5/2.5/111, 8/0.5/2.5/110}
		{
		  \node[label={\scriptsize \label}] (\nodename) at (\x,\y) [nodeDecorate] {};
		}
		\foreach \nodename/\x/\y in {5/0.5/0.5}
		{
		  \node[] (\nodename) at (\x,\y) [nodeDecorate] {};
		}
		\path \foreach \startnode/\endnode in {
		  1/2, 2/3, 3/4, 4/1, 6/7, 7/8, 2/6, 3/7, 4/8}
		{
		  (\startnode) edge[-,thick] node {} (\endnode)
		};
		
		\path \foreach \startnode/\endnode in {5/6, 8/5, 1/5}
		{
		  (\startnode) edge[-,thin] node {} (\endnode)
		};
		
		\path (3.5,1.25) edge[double,thick,-latex] node[above,midway] {\scriptsize induces} (4.75,1.25);

		\begin{scope}[xshift=7.75cm]
		\node[anchor=east] at (0.25,1.25) {$q(x)=(f \circ \phi^{-1}) \Biggl($};
		\node[anchor=west] at (5.75,1.25) {$\Biggr)$};
		\node at (3,-1) {Distribution over fixed-point numbers $x$}; 
		
		\node (n0) at (3,3) [] {\scriptsize$[0,1)$};

		\node (n00) at (1.5,2) [] {\scriptsize$[0,\frac{1}{2})$};
		\node (n01) at (4.5,2) [] {\scriptsize$[\frac{1}{2},1)$};

		\node (n000) at (0.75,1) [] {\scriptsize$[0,\frac{1}{4})$};
		\node (n001) at (2.25,1) [] {\scriptsize$[\frac{1}{4},\frac{2}{4})$};
		\node (n010) at (3.75,1) [] {\scriptsize$[\frac{2}{4},\frac{3}{4})$};
		\node (n011) at (5.25,1) [] {\scriptsize$[\frac{3}{4},1)$};

		\node (n0000) at (0.75-0.36,0) [] {\scriptsize$[0,\frac{1}{8})$};
		\node (n0001) at (0.75+0.36,0) [] {\scriptsize$[\frac{1}{8},\frac{2}{8})$};
		\node (n0010) at (2.25-0.36,0) [] {\scriptsize$[\frac{2}{8},\frac{3}{8})$};
		\node (n0011) at (2.25+0.36,0) [] {\scriptsize$[\frac{3}{8},\frac{4}{8})$};
		\node (n0100) at (3.75-0.36,0) [] {\scriptsize$[\frac{4}{8},\frac{5}{8})$};
		\node (n0101) at (3.75+0.36,0) [] {\scriptsize$[\frac{5}{8},\frac{6}{8})$};
		\node (n0110) at (5.25-0.36,0) [] {\scriptsize$[\frac{6}{8},\frac{7}{8})$};
		\node (n0111) at (5.25+0.36,0) [] {\scriptsize$[\frac{7}{8},1)$};

		\path (n0) edge[->] node[right,midway] {\scriptsize$b_1$} (n01);
		\path (n0) edge[->] node[left] {\scriptsize$\neg b_1$} (n00);
		
		\path (n00) edge[->] node[left] {\scriptsize$\neg b_2$} (n000);
		\path (n00) edge[->] node[right] {\scriptsize$b_2$} (n001);
		\path (n01) edge[->] node[left] {\scriptsize$\neg b_2$} (n010);
		\path (n01) edge[->] node[right] {\scriptsize$b_2$} (n011);

		\path (n000) edge[->] node[left] {\scriptsize$\neg b_3$} (n0000);
		\path (n000) edge[->] node[right] {\scriptsize$b_3$} (n0001);
		\path (n001) edge[->] node[left] {\scriptsize$\neg b_3$} (n0010);
		\path (n001) edge[->] node[right] {\scriptsize$b_3$} (n0011);
		\path (n011) edge[->] node[left] {\scriptsize$\neg b_3$} (n0110);
		\path (n011) edge[->] node[right] {\scriptsize$b_3$} (n0111);
		\path (n010) edge[->] node[left] {\scriptsize$\neg b_3$} (n0100);
		\path (n010) edge[->] node[right] {\scriptsize$b_3$} (n0101);
		
		\end{scope}

	\end{tikzpicture}
	\caption{\textbf{Illustration of our method:} For the case of fixed-point numbers, we use the bitstring to up to each sum node to index the sum in the circuit. The bitstring can be visualized as a hypercube, and the PC induces a distribution over the fixed-point numbers represented by the bitstring.}
	\label{fig:method}
\end{figure*}
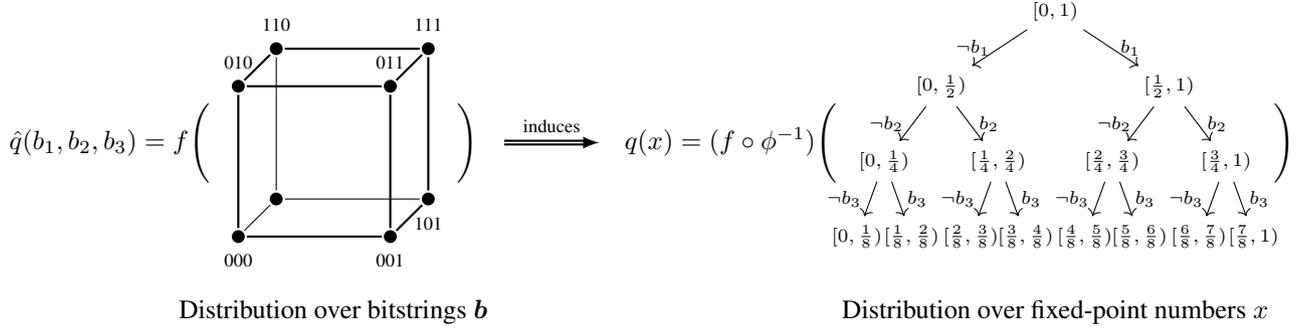

\section{Background and Related Work}
\label{sec:related-work}

The relationship between continuous and discrete representations is fundamental to computational science. At its core, digital computation relies on discrete structures, with real-valued quantities encoded as finite-length bitstrings \citep[Ch.~4][]{knuth2014art}. Floating-point arithmetic provides an approximation to continuous values within this discrete framework, ensuring efficient numerical operations while introducing inherent precision limitations \citep[Ch.~1][]{sterbenz1974floating}. In recent years, this foundational connection has gained renewed attention in machine learning, particularly due to advances in quantization and low-precision arithmetic. 
While these techniques are primarily motivated by hardware constraints, they also present an opportunity: if inference can be formulated directly over discrete bitstring representations, it may unlock new efficiencies in probabilistic modeling.

Bayesian inference provides a principled framework for reasoning under uncertainty, yet exact inference remains intractable in most real-world scenarios. This has led to the development of approximate inference techniques, such as variational inference (VI) \citep{blei2017variational,jordan1999introduction,wainwright2008graphical}. VI formulates inference as an optimization problem, where a parametric distribution is fitted to approximate the posterior while minimizing the reverse KL divergence. Despite its scalability, VI is often constrained by its reliance on continuous parameterizations, which can introduce numerical instabilities and bias due to restrictive approximations, \eg, mean-field or unimodality assumptions. These limitations are apparent when operating under low-precision, raising the question: \emph{Can we perform inference directly in a discrete representation space?}\looseness-1

Probabilistic circuits (PCs) are a recent framework to study tractable representations of complex probability distributions \citep{choi2020probabilistic}.
Depending on the structural properties of the PC, certain inference scenarios can be rendered tractable (polynomial in the model complexity) under the circuit while maintaining a high expressivity.
While PCs are typically employed for exact probabilistic inference, they have found successful application in approximate Bayesian inference, for example, as surrogate through compilation \citep{lowd2010approximate}, as variational distribution for structured discrete models \citep{shih2020probabilistic}, or in discrete probabilistic programs \citep{saad2021sppl}. 
Our work is closely related to work by \citet{garg2024bit}, which utilized PCs over bitstring representation for efficient approximate inference in probabilistic programs.
This work highlights that PCs are a promising representational framework for approximate Bayesian and uncertainty quantification.

\section{Methods}
\label{sec:methods}
Given a target density $p$, we aim to find a variational approximation $q$ that minimizes the divergence of $p$ from $q$. As commonly done, we will focus on the reverse  Kullback–Leibler (KL) divergence of $q$ from $p$, instead of the forward KL. Moreover, we assume that $q$ takes a parametric form with parameters $\btheta$, \ie, $q_{\btheta}$.
Thus, the goal is to find $\btheta$ such that 
\begin{equation}
  \KL(q_{\btheta} \,\|\, p) = \int_{x \in \mcX} q_{\btheta}(x) \log \left( \frac{q_{\btheta}(x)}{p(x)} \right) \dee x , \label{eq:reverse-kl}
\end{equation}
is minimized, assuming that $\mcX \subseteq \reals^d$ for some $d\geq 1$.

In general, computing \cref{eq:reverse-kl} is intractable for two reasons: {\em (i)}~$p$ is often only known up to an unknown normalization constant $Z_p$ and {\em (ii)}~$p$ and $q$ do not exhibit sufficient structure to render the integration tractable \citep{wang2024atlas}.
Henceforth, one typically optimizes the evidence lower bound (ELBO), which can be written as
\begin{equation}
  \mcL(q_{\btheta}, p) = \EE_{x \sim q_{\btheta}}\left[\log p(x)\right] +\ent{q_{\btheta}}, \label{eq:elbo}
\end{equation}
where $\ent{q_{\btheta}} = -\EE_{x \sim q_{\btheta}}\left[\log q_{\btheta}(x)\right]$ denotes the entropy of the variational distribution $q_{\btheta}$. In case $q_{\btheta}$ admits a tractable entropy computation, only the first term in \cref{eq:elbo} requires numerical approximation.

When computing either \cref{eq:reverse-kl} or \cref{eq:elbo} on a computer, each $x$ will inevitably be represented in a discretized form. In fact, every real-valued number is represented by a series of bitstrings and mapped to the real line by a mapping function $\phi\colon \{0,1\}^B \to \reals$ given by the chosen number system. 
Consequently, any distribution $p$ or $q$ represented on a computer can be expressed in terms of a distribution over bitstrings.

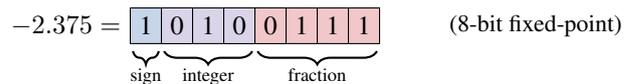
\begin{figure}[h!]
	\centering	
	\begin{tikzpicture}[
	  node distance=0pt,
	  start chain = A going right,
	  X/.style = {rectangle, draw,%
	                minimum width=2ex, minimum height=3ex,
	                outer sep=0pt, on chain},
	  B/.style = {decorate,
	                decoration={brace, amplitude=5pt,
	                pre=moveto,pre length=1pt,post=moveto,post length=1pt,
	                raise=1mm,
	                            #1}, %
	                thick},
	  B/.default=mirror,
	  ]
	  \node[minimum width=8ex, on chain] {$-2.375 = $};
	  \foreach \i in {1}%
	      \node[X, fill=colour1!30] (s) {\i};
	      
	  \foreach \i/\b in {1/0,2/1,3/0} %
	      \node[X, fill=colour4!30] (i\i) {\b};
	        
	  \foreach \i/\b in {1/0,2/1,3/1,4/1} %
	      \node[X, fill=colour3!30] (f\i) {\b};
	      
	  \draw[B] (s.south west) -- node[below=2mm] {\scriptsize sign} (s.south east);
	  \draw[B] (i1.south west) -- node[below=2mm] {\scriptsize integer} (i3.south east);
	  \draw[B] (f1.south west) -- node[below=2mm] {\scriptsize fraction} (f4.south east);  
	  
	  \node[right of=f4, anchor=west, xshift=1cm] {\small (8-bit fixed-point)};
	\end{tikzpicture}
	\caption{Representation of `$-2.375$' using an 8-bit fixed-point number system with sign, integer, and fraction bits.}
	\label{fig:fixedpoint}
\end{figure}

\textbf{Fixed-point representations} In this work, we focus on the fixed-point number representation system for bitstrings, which interprets a $B$-bit bitstring as containing one sign bit, $F$ fractional bits and $B - F - 1$ integer bits. An example of an 8-bit fixed-point bitstring is illustrated in \cref{fig:fixedpoint}, and its mapping function $\phi$ is defined in \cref{app:fp-mapping}.

The fixed-point representation was chosen for this work, as its mapping function iteratively partitions an interval (defined by the smallest and largest value $\phi$ can represent) in $\reals$ into fixed-width sub-intervals. As such, this representation is easier to understand and visualize. It is important to note that our approach, however, is not strictly limited to fixed-point representations. For example, the more commonly used floating-point bitstrings can also be used with our method as this representation mainly differs in using variable-width sub-intervals.

In the following, we outline how a continuous distribution can be approximated with a tractable and flexible variational family defined over its fixed-point bitstring representation.

\subsection{BitVI: Variational Distributions over Bitstring Representations}
\label{sec:BitVI}
Let $\hat{q}$ be a distribution over binary strings with probability measure $\hat{Q}$ defined on the measurable space of binary strings $(\mcY, \mcA)$ with corresponding $\sigma$-algebra $\mcA$.
Further, let $(\reals, \mcB)$ be the measurable space of real numbers with Borel $\sigma$-algebra $\mcB$.
Define a measurable mapping $\phi\colon \mcY \to \reals$ that assigns to each binary string a real number according to a specified number system, for example, the fixed point representation.
The induced probability measure $Q$ on $(\reals, \mcB)$ is the pushforward measure of $\hat{Q}$ through $\phi$. 
Specifically, for any Borel set $B \in \mcB$ we have $Q(B) = \hat{Q}(\phi^{-1}(B))$ where $\phi^{-1}(B)$ is the pre-image of $B$ under $\phi$.
Finally, we represent the density $q$ of $Q$ using a (deterministic) probabilistic circuit (PC).
This construction is illustrated in \cref{fig:method} for the case of fixed-point numbers, where we use the bitstring up to each sum node to index the sum in the circuit.
For fixed-point representations with infinite precision, this construction is equivalent to probability measures generated by P\'olya trees \citep{ferguson1974polya,trapp2022priors}.

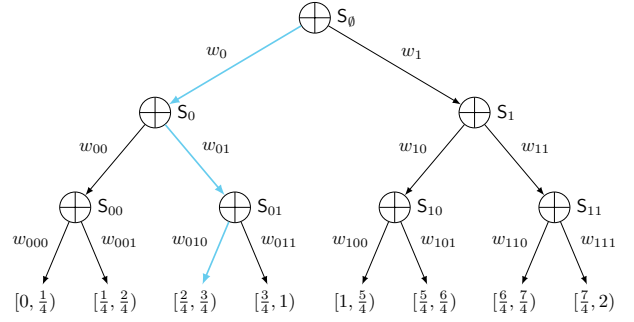
\begin{figure}
  \centering
  \resizebox{\columnwidth}{!}{
  \begin{tikzpicture}[
    ->,
    >=latex,
    level 1/.style={sibling distance=6cm, level distance=1.8cm},
    level 2/.style={sibling distance=3cm, level distance=1.8cm},
    level 3/.style={sibling distance=1.5cm, level distance=1.8cm},
    align=center,
    selection/.style={edge from parent/.style={thick,draw=scCyan}},
    noselection/.style={edge from parent/.style={draw,black,thin}},
    emphs/.style={edge from parent/.style={thin,dashed,draw,black}, level distance = 1cm, sibling distance = 1cm},
    cross/.style={-,path picture={ \draw[black]
  (path picture bounding box.east) -- (path picture bounding box.west) (path picture bounding box.south) -- (path picture bounding box.north);
  }},
  scale=0.9,
  transform shape
    ]
      \node[draw, circle, cross, text width=3mm, label={right:$\snode_{\emptyset}$}] at (0,0) {}
        child[selection]{ node[draw, circle, cross, text width=3mm, label={right:$\snode_{0}$}] {}
          child[noselection]{ node[draw,black,thin,circle, cross, text width=3mm, label={right:$\snode_{00}$}] {} 
              child[noselection]{ node[] {$[0, \frac{1}{4})$} 
                  edge from parent node[above left] {$w_{000}$}
              }
              child[noselection]{ node[] {$[\frac{1}{4}, \frac{2}{4})$} 
                  edge from parent node[above right] {$w_{001}$}
              }
              edge from parent node[above left] {$w_{00}$}
          }
          child[selection]{ node[draw,black,thin, circle, cross, text width=3mm, label={right:$\snode_{01}$}] {} 
              child[selection]{ node[] {$[\frac{2}{4}, \frac{3}{4})$} 
                  edge from parent node[above left] {$w_{010}$}
              }
              child[noselection]{ node[] {$[\frac{3}{4}, 1)$} 
                  edge from parent node[above right] {$w_{011}$}
              }
              edge from parent node[above right] {$w_{01}$}
          }
          edge from parent node[above left] {$w_{0}$}
        }
        child[sibling distance = 6cm]{ node [draw, circle, cross, text width=3mm, label={right:$\snode_{1}$}] {}
            child{ node[draw, circle, cross, text width=3mm, label={right:$\snode_{10}$}] {} 
              child{ node[] {$[1, \frac{5}{4})$} 
                  edge from parent node[above left] {$w_{100}$}
              }
              child[]{ node[] {$[\frac{5}{4}, \frac{6}{4})$} 
                  edge from parent node[above right] {$w_{101}$}
              }
              edge from parent node[above left] {$w_{10}$}
            }
            child{ node[draw, circle, cross, text width=3mm, label={right:$\snode_{11}$}] {} 
              child{ node[] {$[\frac{6}{4}, \frac{7}{4})$} 
                  edge from parent node[above left] {$w_{110}$}
              }
              child[]{ node[] {$[\frac{7}{4}, 2)$} 
                  edge from parent node[above right] {$w_{111}$}
              }
              edge from parent node[above right] {$w_{11}$}
          }
        edge from parent node[above right] {$w_{1}$}
      }
    ;
  \end{tikzpicture}
  }
  \caption{A deterministic PC over a 3-bit fixed-point bitstring. $\snode_\emptyset$ encodes the distribution of the full bitstring, and $\snode_\epsilon$ encodes the distribution of the remaining sub-bitstring given $\epsilon$. Intervals $[a, b)$ denote the leaf nodes of the circuit, which are uniform distributions.} 
  \label{fig:bitpc}
\end{figure}

\begin{definition}[Deterministic Probabilistic Circuit]
A probabilistic circuit $\pc(\bx)$ is a multi-linear function represented by a computational graph consisting of three types of nodes $\node(\bx)$; Sum nodes $\snode(\bx) = \sum_i w_i \cnode_i(\bx)$ and product nodes $\pnode(\bx) = \prod_i \cnode_i(\bx)$ which compute a function of their child nodes $\cnode_i(\bx)$ respectively, and leaf nodes consisting of tractable (univariate) functions $\lnode(x) = p(x)$.
The circuit $\pc$ characterizes a multivariate probability distribution over random variables $\mcX = \{X_1, \dots, X_d\}$ by, for example, representing its mass, density, or characteristic function \citep{yu2023cc,broadrick2024semantics}.
Note that we assume that the circuit is smooth and decomposable \citep{choi2020probabilistic} and refer to \cref{app:details} for details. 

A sum node $\snode$ is deterministic if for all $\bx$, only one summand is non-zero. 
Consequently, $\pc$ is deterministic if all sum nodes are deterministic \citep{choi2020probabilistic}.
\end{definition}
By specifying a $\hat{q}$ over bitstrings and a respective number system, we obtain an induced variational distribution $q$ on the real line.
As previously mentioned, our goal is to find a parameterization $\btheta$ of our variational distribution such that \cref{eq:reverse-kl} is minimal.
When representing $q$ using a deterministic PC, the parameters $\btheta$ correspond to the collection of weights $\{w_i\}_i$ of the circuit.
Note that by construction, the leaf nodes of our circuit model are continuous uniform distributions and, therefore, do not have any additional parameters.
The resulting deterministic PC is a tree with depth proportional to the number of bits used in the bitstring representation.
Each sum node in the PC represents the decision of a bit and weights correspond to the conditional probability of the respective decision.
For example, the probability of $0.5$ in 3-bit fixed-point number system with one integer bit and no sign-bit, which corresponds to the bitstring $010$, is computed by obtaining the bit decisions, \ie, $b_0=0$, $b_1=1$, and $b_2=0$, and evaluating the circuit along the respective path, \ie, $p(x = 0.5) = w_{0} w_{01} w_{010} \frac{1}{2^{B_{\text{frac}}}}$ where $B_{\text{frac}} = 2$ is the number of fraction bits. \cref{fig:bitpc} illustrates the decision process represented by the circuit.

\paragraph{Depth Regularization}
To encourage that $q$ has a smooth density in the limit of infinite precision, we leverage a depth regularization. The depth regularization is based on P\'olya tree prior constructions for priors over continuous probability distributions.
Specifically, \citet{ferguson1974polya} proposed to use a Beta prior on each weight of a P\'olya tree with symmetric $\alpha$-parameter that has a quadratic increase in depth $j$ of the tree, \ie, $\alpha(j)= j^2$.
An alternative parameterization is given by \citet{castillo2017polya} as $\alpha(j)= 2^{j}$.
Both approaches ensure that the prior probability of uniformly distributed weights increases with depth.
We adopt this approach and use Laplace smoothing of the circuit weights with a depth-dependent smoothing factor.
In particular, for bit $b_j$ (depth $j$) with $j \geq 0$ we define each weight for $b_j=0$ as
\begin{equation}
  w_{\epsilon0} = \frac{v_{\epsilon0} + c\alpha(j) }{ v_{\epsilon0} + v_{\epsilon1} + 2c\alpha(j) } ,
\end{equation}
where $\epsilon$ denotes a $j-1$ long binary string, $v_{\epsilon0} > 0$ is an unnormalized weight, and $c > 0$ is a hyperparameter. The weight for $\epsilon1$ is given analogously.

\begin{figure*}
  \centering
  \def\datapath{.}
  \setlength{\figurewidth}{.19\textwidth}
  \setlength{\figureheight}{\figurewidth}
  \pgfplotsset{
    scale only axis,
    axis on top,
    ytick=\empty,        %
    yticklabels=\empty,  %
    xtick=\empty,        %
    xticklabels=\empty,  %
  }
  \begin{subfigure}[t]{.19\textwidth}
    \centering
    \input{mle_moons.tex}\\[-.5em]
	\caption{Deterministic}
  \end{subfigure}
  \hfill
  \begin{subfigure}[t]{.19\textwidth}
    \centering
    \input{ffg_moons.tex}\\[-.5em]
	\caption{MFVI}
  \end{subfigure}
  \hfill
  \begin{subfigure}[t]{.19\textwidth}
    \centering
    \input{2bitvi_moons.tex}\\[-.5em]
	\caption{\ours\ (2 bits)}
  \end{subfigure}
  \hfill
  \begin{subfigure}[t]{.19\textwidth}
    \centering
    \input{4bitvi_moons.tex}\\[-.5em]
	\caption{\ours\ (4 bits)}
  \end{subfigure}
  \hfill
  \begin{subfigure}[t]{.19\textwidth}
    \centering
    \input{8bitvi_moons.tex}\\[-.5em]
	\caption{\ours\ (8 bits)}
  \end{subfigure}
  \newcommand{\cmap}{\protect\includegraphics[width=2.5em,height=.6em]{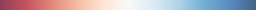}}
  \newcommand{\mylegend}{\protect\tikz[baseline=-.6ex, opacity=0.8]{\protect\draw[white, fill=red] (0, 0) rectangle ++(4pt,4pt);\protect\draw [white, fill=blue] (4pt, 0) circle (2pt);}}
  \caption{\textbf{Uncertainty quantification in neural networks:} We consider the two moons binary \mylegend\ classification problem with an MLP neural network (two hidden layers). The predictive density (\cmap) shows that BitVI provides both representative uncertainties and good decision boundaries compared to the deterministic and MFVI baselines.}
  \label{fig:twomoons}
\end{figure*}

\paragraph{Computation of the ELBO}
A particular property of deterministic PCs is that the entropy can be computed in linear time w.r.t.\ the number of edges of the circuit \citep{vergari2021compositional} (see \cref{app:derivations} for details).
As such, we only need to approximate the expected log probability in \cref{eq:elbo} using Monte Carlo (MC) integration.
To do so, we first use a reparameterization using the inverse CDF transform, which is available analytically in the case of deterministic PCs.

In particular, we reparameterize the ELBO as, 
\begin{equation}
    \mcL(q_{\btheta}, p) = \EE_{u \sim \distUnif(0,1)}\left[\log p(F^{-1}_{q_{\btheta}}(u))\right] +\ent{q_{\btheta}}, \label{eq:reparam_elbo}
\end{equation}
where $F^{-1}_{q_{\btheta}}(\cdot)$ is the inverse CDF transform of $q_{\btheta}$.
We then generate $T$ samples from a uniform distribution $u^s\sim\distUnif(0,1)$ and compute a MC estimate of \cref{eq:reparam_elbo}, \ie,
\begin{equation}
    \mcL(q_{\btheta}, p) \approx \frac{1}{T}\sum^S_{s=1} \log p(F^{-1}_{q_{\btheta}}(u_s)) +\ent{q_{\btheta}}. \label{eq:reparam_elbo_mc}
\end{equation}

Note that \cref{eq:reparam_elbo_mc} can be computed efficiently.
\begin{remark}
The inverse CDF transform of $q_{\btheta}$ can be computed in linear time w.r.t.\ the depth of the circuit.
\end{remark}

For a given input $y$, we can compute the inverse CDF transform of $y$ under $q_{\btheta}$ using a series of linear transformations.
In particular, sum nodes $\snode$ compute
\begin{equation}
  F^{-1}_{\snode_{\epsilon}}(y) = \begin{cases}
    F^{-1}_{\cnode_{\epsilon1}}\left(\frac{y - w_{\epsilon0}}{w_{\epsilon1}}\right) & \text{if } y > w_{\epsilon0} \\
    F^{-1}_{\cnode_{\epsilon0}}\left(\frac{y}{w_{\epsilon0}}\right) & \text{otherwise }
  \end{cases} , \label{eq:local_moves}
\end{equation}
where $\cnode$ denotes a child node of $\snode$, \ie a sum or leaf node, and $\epsilon \in \bigcup^B_{j=0} \{0,1\}^j$ is a bitstring.
If $\cnode$ is a leaf node, we compute the inverse CDF according to the leaf distribution, \ie, $F^{-1}_{\lnode}(y) = y(b-a) + a$ in case of a continuous uniform distribution $\mcU(a,b)$. See further details in \cref{app:derivations}.

Note that the resulting value still requires discretization, and in case of fixed-point numbers needs to be rounded to the nearest fixed-point value.
In fact, the bitstring $\epsilon$ generated by traversing the circuit in order to compute its inverse CDF already encodes the nearest fixed-point value for $y$.
However, as the discretization operation does not have a well-defined gradient, we resort to the application of the straight-through estimator (STE) \citep{bengio2013estimating}.
In particular, we compute:
\begin{equation}
  x =  (\phi(\epsilon) + F^{-1}_{q_{\btheta}}(y)) - F^{-1}_{q_{\btheta}}(y) ,
\end{equation}
where $\phi(\epsilon)$ is the mapping function defined by the number system and the bitstring $\epsilon$ is a function of $F^{-1}_{q_{\btheta}}$ and indicates the decision taken in \cref{eq:local_moves}.

\begin{table*}[t]
	\centering\scriptsize
	\caption{\textbf{Bayesian benchmarks:} Negative log predictive density (NLPD\scalebox{.75}{\textcolor{black!50}{${\pm}$std}}, smaller better) results on the {\em Bayesian Benchmarks} UCI tasks (5-fold CV). We compare \ours\ to Gaussian MFVI and Full-covariance Gaussian VI (FCGVI) on small MLP NN models. The best-performing method for each task is bolded, and multiple methods are bolded based on a paired $t$-test ($p=5\%$). We show that \ours\ works well on all test cases and is not significantly different from the baselines in most cases, even in the very low-bit range.}
	\label{tbl:uci_nlpd}

    \setlength{\tabcolsep}{12pt}

	\newcommand{\val}[3]{$#1#2$\scalebox{.75}{\textcolor{black!50}{${\pm}#3$}}}  
	
	\begin{tabular}{lc|ccccc}
\hline
Dataset & $(n,d)$ & MFVI & FCGVI & 2-BitVI & 4-BitVI & 8-BitVI\\
\hline
\textsc{fertility} & (100,10) & \val{\bf}{0.379}{0.107} & \val{}{0.406}{0.111} & \val{}{0.728}{0.139} & \val{\bf}{0.407}{0.109} & \val{\bf}{0.406}{0.142}\\
\textsc{pittsburg-bridges-T-OR-D} & (102,8) & \val{\bf}{0.345}{0.168} & \val{\bf}{0.347}{0.078} & \val{\bf}{0.301}{0.064} & \val{\bf}{0.352}{0.082} & \val{\bf}{0.391}{0.068}\\
\textsc{acute-inflammation} & (120,7) & \val{\bf}{0.004}{0.001} & \val{}{0.021}{0.009} & \val{\bf}{0.006}{0.002} & \val{\bf}{0.006}{0.002} & \val{}{0.684}{0.031}\\
\textsc{acute-nephritis} & (120,7) & \val{\bf}{0.003}{0.001} & \val{}{0.014}{0.003} & \val{\bf}{0.002}{0.000} & \val{\bf}{0.002}{0.002} & \val{}{0.051}{0.016}\\
\textsc{echocardiogram} & (131,11) & \val{\bf}{0.446}{0.167} & \val{\bf}{0.515}{0.151} & \val{\bf}{0.524}{0.200} & \val{\bf}{0.435}{0.095} & \val{}{0.660}{0.132}\\
\textsc{hepatitis} & (155,20) & \val{\bf}{0.438}{0.081} & \val{\bf}{0.447}{0.116} & \val{\bf}{0.620}{0.246} & \val{\bf}{0.694}{0.279} & \val{\bf}{0.427}{0.085}\\
\textsc{parkinsons} & (195,23) & \val{\bf}{0.322}{0.151} & \val{\bf}{0.284}{0.109} & \val{\bf}{0.253}{0.098} & \val{\bf}{0.261}{0.064} & \val{\bf}{0.289}{0.061}\\
\textsc{breast-cancer-wisc-prog} & (198,34) & \val{\bf}{0.540}{0.106} & \val{\bf}{0.522}{0.128} & \val{}{0.699}{0.087} & \val{\bf}{0.584}{0.073} & \val{\bf}{0.548}{0.087}\\
\textsc{spect} & (265,23) & \val{\bf}{0.614}{0.067} & \val{\bf}{0.624}{0.053} & \val{}{0.801}{0.108} & \val{}{0.807}{0.148} & \val{\bf}{0.670}{0.125}\\
\textsc{statlog-heart} & (270,14) & \val{\bf}{0.478}{0.133} & \val{\bf}{0.488}{0.156} & \val{\bf}{0.550}{0.207} & \val{\bf}{0.606}{0.270} & \val{\bf}{0.478}{0.147}\\
\textsc{haberman-survival} & (306,4) & \val{\bf}{0.535}{0.062} & \val{\bf}{0.523}{0.054} & \val{\bf}{0.531}{0.042} & \val{\bf}{0.525}{0.044} & \val{\bf}{0.530}{0.036}\\
\textsc{ionosphere} & (351,34) & \val{\bf}{0.288}{0.094} & \val{\bf}{0.276}{0.092} & \val{\bf}{0.335}{0.126} & \val{\bf}{0.459}{0.217} & \val{\bf}{0.323}{0.127}\\
\textsc{horse-colic} & (368,26) & \val{\bf}{0.611}{0.159} & \val{\bf}{0.595}{0.163} & \val{}{0.618}{0.119} & \val{}{0.690}{0.143} & \val{\bf}{0.576}{0.103}\\
\textsc{congressional-voting} & (435,17) & \val{\bf}{0.670}{0.093} & \val{\bf}{0.700}{0.126} & \val{\bf}{0.699}{0.105} & \val{\bf}{0.704}{0.108} & \val{\bf}{0.644}{0.048}\\
\textsc{cylinder-bands} & (512,36) & \val{\bf}{0.602}{0.107} & \val{\bf}{0.633}{0.050} & \val{}{0.835}{0.222} & \val{\bf}{0.955}{0.361} & \val{\bf}{0.678}{0.019}\\
\textsc{breast-cancer-wisc-diag} & (569,31) & \val{\bf}{0.078}{0.050} & \val{\bf}{0.108}{0.029} & \val{\bf}{0.148}{0.080} & \val{\bf}{0.172}{0.152} & \val{\bf}{0.155}{0.097}\\
\textsc{ilpd-indian-liver} & (583,10) & \val{\bf}{0.547}{0.059} & \val{\bf}{0.547}{0.033} & \val{\bf}{0.535}{0.053} & \val{\bf}{0.518}{0.032} & \val{\bf}{0.567}{0.025}\\
\textsc{monks-2} & (601,7) & \val{\bf}{0.083}{0.121} & \val{}{0.607}{0.082} & \val{}{0.563}{0.060} & \val{}{0.656}{0.073} & \val{}{0.666}{0.030}\\
\textsc{credit-approval} & (690,16) & \val{\bf}{0.357}{0.025} & \val{\bf}{0.417}{0.096} & \val{}{0.405}{0.041} & \val{\bf}{0.358}{0.026} & \val{\bf}{0.343}{0.009}\\
\textsc{statlog-australian-credit} & (690,15) & \val{\bf}{0.662}{0.035} & \val{}{0.650}{0.029} & \val{}{0.764}{0.075} & \val{\bf}{0.629}{0.019} & \val{\bf}{0.626}{0.019}\\
\textsc{breast-cancer-wisc} & (699,10) & \val{\bf}{0.091}{0.042} & \val{\bf}{0.105}{0.041} & \val{\bf}{0.171}{0.113} & \val{}{0.168}{0.055} & \val{\bf}{0.122}{0.059}\\
\textsc{blood} & (748,5) & \val{\bf}{0.483}{0.058} & \val{\bf}{0.483}{0.036} & \val{\bf}{0.486}{0.057} & \val{\bf}{0.478}{0.043} & \val{\bf}{0.486}{0.039}\\
\textsc{pima} & (768,9) & \val{}{0.516}{0.045} & \val{\bf}{0.507}{0.042} & \val{\bf}{0.512}{0.039} & \val{\bf}{0.492}{0.031} & \val{\bf}{0.492}{0.042}\\
\textsc{mammographic} & (961,6) & \val{\bf}{0.428}{0.039} & \val{}{0.468}{0.044} & \val{\bf}{0.430}{0.053} & \val{\bf}{0.417}{0.039} & \val{\bf}{0.423}{0.049}\\
\textsc{statlog-german-credit} & (1000,25) & \val{\bf}{0.547}{0.066} & \val{\bf}{0.557}{0.086} & \val{}{0.651}{0.092} & \val{\bf}{0.646}{0.101} & \val{}{0.894}{0.249}\\
\hline
\end{tabular}
\end{table*}

\paragraph{Representing Multivariate Distributions}
So far, our induced variational distribution is only defined on the real line (univariate case).
To extend the approach to the multivariate case, we considered two approaches: {\em (i)}~a mean-field variational family, and {\em (ii)}~a variational family model with dependencies between dimensions.
To represent dependencies between the dimensions, we construct a deterministic PC representing the joint distribution over the bits of all the dimensions.
In the case of fixed-point number systems, the resulting circuit model recursively splits the domain into hyper-rectangles by performing axis-aligned splits that alternate between dimensions in the construction.
Note that this construction results in a binary tree consisting of $2^{B*D}$ leaves, where $B$ is the number of bits and $D$ is the number of dimensions.
Thus, making it useful in low-dimensional or low-precision settings.
However, including conditional independencies in the model can result in substantially more compact representations \citep{peharz2020einsum,garg2024bit}.
Further details are provided in \cref{app:mv-details}.

Applying the inverse CDF reparameterization for multivariate densities modeled with \ours\ requires further considerations.
In the case of the mean-field approximation, we apply the inverse CDF reparameterization (described above) independently for each dimension.
If \ours\ represents a variational distribution that models dependencies between dimensions, we employ the inverse of the tree-CDF transformation \citep{awaya2024unsupervised}, which is a map $\reals^D \to [0,1]^D$ where $D$ is the number of dimensions.
In particular, for a given input $\by \in [0,1]^D$, we compute the inverse tree-CDF transform of $\by$ by applying the following axis-aligned linear transformations at each sum node, where $\snode_{d,\epsilon_d}$ denotes the sum node for dimension $d \leq D$ under bitstring $\epsilon_d$.
The axis-aligned transformations are given as:
\begin{equation}
  F^{-1}_{\snode_{d,\epsilon_d}}(y_d) = \begin{cases}
    F^{-1}_{\cnode_{1}}\left(\frac{y_d - w_{d,\epsilon_d 0}}{w_{d,\epsilon_d 1}}\right) & \text{if } y_d > w_{d,\epsilon_d 0} \\
    F^{-1}_{\cnode_{0}}\left(\frac{y_d}{w_{d,\epsilon_d 0}}\right) & \text{otherwise }
  \end{cases} , \label{eq:local_moves}
\end{equation}
where with some abuse of notation $\cnode_{0}$ denotes the left child of $\snode_{d,\epsilon_d}$, which corresponds to a bit value of zero, and $\cnode_{1}$ denotes the right child (bit value of one).
As we alternate dimensions at each level in the tree, decisions are made only based on the `selected' dimension at each step.
Computing the inverse of the tree-CDF transformation can still be performed efficiently, \ie, in $\mcO(B*D)$ for $B$ bits.

\section{Experiments}
\label{sec:experiments}
Our experiments are designed to systematically validate the effectiveness of \ours\ in performing approximate probabilistic inference over bitstring representations. In \cref{sec:2d-densities}, we begin with 2D density estimation to demonstrate the expressiveness of our method in capturing complex non-Gaussian distributions.  
In \cref{sec:mlp-exp}, we explore Bayesian deep learning applications by applying \ours\ to MLP neural networks (NNs), showcasing its ability to perform effective uncertainty quantification in predictive modeling. We then conduct a series of ablation studies in \cref{sec:ablations} to assess the trade-offs between numerical precision and model expressivity, investigating the effect of bitstring depth on performance and the role of hierarchical structure in NNs.

\paragraph{Implementation} The method was implemented in Python using the PyTorch library in order to facilitate automatic differentiation, convenient construction of neural network architectures, and fast parallelized training on GPUs. The training was conducted on a high-performance computing cluster with NVIDIA [H,A,V,P]100, K80, and H200 GPUs. As a ballpark, the model training run time for single models in the experiments is measured in the range of minutes for the size of models we consider in these experiments.

\subsection{2D Densities}
\label{sec:2d-densities}
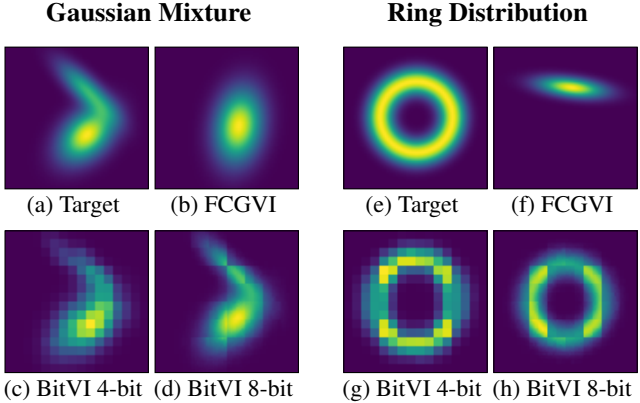
\begin{figure}[t!]
	\centering
	\def\datapath{.}
	\setlength{\figurewidth}{.11\textwidth}
	\setlength{\figureheight}{\figurewidth}
	\pgfplotsset{
	  scale only axis,
	  axis on top,
	  ytick=\empty,        %
	  yticklabels=\empty,  %
	  xtick=\empty,        %
	  xticklabels=\empty,  %
	}
		
	\begin{minipage}[t]{0.46\columnwidth}
	 \centering
	  {\bf\strut Gaussian Mixture}\\[.5em]
	  \begin{subfigure}[t]{.48\textwidth}
	  \centering
\begin{tikzpicture}

\begin{axis}[
height=\figureheight,
tick pos=left,
width=\figurewidth,
xmin=-4, xmax=3.90000000000001,
ymin=-4, ymax=3.90000000000001
]
\addplot graphics [includegraphics cmd=\pgfimage,xmin=-4, xmax=3.90000000000001, ymin=-4, ymax=3.90000000000001] {true_mixture-000.png};
\end{axis}

\end{tikzpicture}\\[-.5em]
      \caption{Target}
	\end{subfigure}
    \hfill
	\begin{subfigure}[t]{.48\textwidth}
	  \centering
\begin{tikzpicture}

\begin{axis}[
height=\figureheight,
tick pos=left,
width=\figurewidth,
xmin=-4, xmax=3.90000000000001,
ymin=-4, ymax=3.90000000000001
]
\addplot graphics [includegraphics cmd=\pgfimage,xmin=-4, xmax=3.90000000000001, ymin=-4, ymax=3.90000000000001] {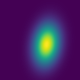};
\end{axis}

\end{tikzpicture}\\[-.5em]
	  \caption{FCGVI}
	\end{subfigure}\\[.25em]
	\begin{subfigure}[t]{.48\textwidth}
	  \centering
\begin{tikzpicture}

\begin{axis}[
height=\figureheight,
tick pos=left,
width=\figurewidth,
xmin=-4, xmax=3.90000000000001,
ymin=-4, ymax=3.90000000000001
]
\addplot graphics [includegraphics cmd=\pgfimage,xmin=-4, xmax=3.90000000000001, ymin=-4, ymax=3.90000000000001] {bitvi_mixture_4bit-000.png};
\end{axis}

\end{tikzpicture}\\[-.5em]
      \caption{\ours\ 4-bit}
	\end{subfigure}
    \hfill
	\begin{subfigure}[t]{.48\textwidth}
	  \centering
\begin{tikzpicture}

\begin{axis}[
height=\figureheight,
tick pos=left,
width=\figurewidth,
xmin=-4, xmax=3.90000000000001,
ymin=-4, ymax=3.90000000000001
]
\addplot graphics [includegraphics cmd=\pgfimage,xmin=-4, xmax=3.90000000000001, ymin=-4, ymax=3.90000000000001] {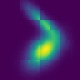};
\end{axis}

\end{tikzpicture}\\[-.5em]
	  \caption{\ours\ 8-bit}
	\end{subfigure}
	\end{minipage}
	\hfill
	\begin{minipage}[t]{0.46\columnwidth}
	\centering
	  {\bf\strut Ring Distribution}\\[.5em]
	  \begin{subfigure}[t]{.48\textwidth}
	  \centering
\begin{tikzpicture}

\begin{axis}[
height=\figureheight,
tick pos=left,
width=\figurewidth,
xmin=-4, xmax=3.90000000000001,
ymin=-4, ymax=3.90000000000001
]
\addplot graphics [includegraphics cmd=\pgfimage,xmin=-4, xmax=3.90000000000001, ymin=-4, ymax=3.90000000000001] {true_ring-000.png};
\end{axis}

\end{tikzpicture}\\[-.5em]
      \caption{Target}
	\end{subfigure}
    \hfill
	\begin{subfigure}[t]{.48\textwidth}
	  \centering
\begin{tikzpicture}

\begin{axis}[
height=\figureheight,
tick pos=left,
width=\figurewidth,
xmin=-4, xmax=3.90000000000001,
ymin=-4, ymax=3.90000000000001
]
\addplot graphics [includegraphics cmd=\pgfimage,xmin=-4, xmax=3.90000000000001, ymin=-4, ymax=3.90000000000001] {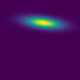};
\end{axis}

\end{tikzpicture}\\[-.5em]
	  \caption{FCGVI}
	\end{subfigure}\\[.25em]
	\begin{subfigure}[t]{.48\textwidth}
	  \centering
\begin{tikzpicture}

\begin{axis}[
height=\figureheight,
tick pos=left,
width=\figurewidth,
xmin=-4, xmax=3.90000000000001,
ymin=-4, ymax=3.90000000000001
]
\addplot graphics [includegraphics cmd=\pgfimage,xmin=-4, xmax=3.90000000000001, ymin=-4, ymax=3.90000000000001] {bitvi_ring_4bit-000.png};
\end{axis}

\end{tikzpicture}\\[-.5em]
      \caption{\ours\ 4-bit}
	\end{subfigure}
    \hfill
	\begin{subfigure}[t]{.48\textwidth}
	  \centering
\begin{tikzpicture}

\begin{axis}[
height=\figureheight,
tick pos=left,
width=\figurewidth,
xmin=-4, xmax=3.90000000000001,
ymin=-4, ymax=3.90000000000001
]
\addplot graphics [includegraphics cmd=\pgfimage,xmin=-4, xmax=3.90000000000001, ymin=-4, ymax=3.90000000000001] {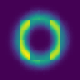};
\end{axis}

\end{tikzpicture}\\[-.5em]
	  \caption{\ours\ 8-bit}
	\end{subfigure}
	\end{minipage}
		
	\caption{Comparison of 4-bit/8-bit \ours\ against full-covariance Gaussian VI (FCGVI) on 2D non-Gaussian target distributions. A full comparison on all target distributions is given in \cref{fig:densities-app}. \ours\ captures the overall density and cross-dependencies better than FCGVI.}
	\label{fig:densities-comparison}
\end{figure}

First, we demonstrate the flexibility of our proposed approach in 2D non-Gaussian target distributions. In \cref{fig:2d_densities}, we include typical benchmark target densities (mixture, Neal's funnel, two-modal Gaussian, ring, and banana) that we approximate with 4-bit \ours\ . Moreover, \cref{fig:densities-comparison} shows a comparison for two densities, indicating that \ours\ captures the overall density and cross-dependencies well, with approximation quality increasing with the number of bits. \cref{fig:densities-app} in the Appendix shows comparisons to the remaining densities. 

\subsection{MLP Neural Network Models}
\label{sec:mlp-exp}
We experiment with probabilistic inference in multi-layer perceptron (MLP) neural network (NN) models. For simplicity, we use similar neural network architectures in all the NN experiments. We use two hidden layers in all experiments, only varying the number of units. Additionally, we use the layer norm to ensure weight scaling. 

\cref{fig:twomoons} shows an uncertainty quantification example. We consider the two moons binary classification problem with an MLP neural network ([8,8] hidden units). The predictive density shows that \ours\ provides both representative uncertainties and good decision boundaries compared to the deterministic and mean-field Gaussian VI baselines.

To give a more quantitative treatment to MLP NN modeling tasks, we use the {\em Bayesian Benchmarks}\footnote{\tiny \url{github.com/secondmind-labs/bayesian_benchmarks}; originally by Salimbeni~\etal} community suite meant for benchmarking Bayesian methods in machine learning. Bayesian benchmarks include common evaluation data sets (typically from UCI \citep{UCI}) and make it possible to run a large number of comparisons under a fixed evaluation setup. We evaluate our approach in binary classification, and for an interesting probabilistic treatment, we include small-data binary classification tasks with $100 \leq n \leq 1000$ data samples (25 data sets). We follow the standard setup of input point normalization and splits in the evaluation suite. Additional details on the NN architectures and evaluation setup can be found in \cref{app:exp-MLP}.

\cref{tbl:uci_nlpd} shows the results for \ours\ (with 2, 4, and 8 bits), mean-field Gaussian VI (MFVI), and full-covariance Gaussian VI (FCGVI). Our approach consistently performs competitively with the standard variational inference baselines, even in the low-bit regime. Notably, in most data sets, \ours\ with 4-bit and 8-bit representations achieves comparable performance to MFVI and FCGVI, demonstrating that probabilistic inference can be effectively conducted over bitstring representations without significant loss in predictive power. Even at 2-bit precision, \ours\ remains viable in several cases. Yet, the results also suggest that more flexible probabilistic modeling in this neural network setting might not be needed, as the 8-bit models show very little or any benefits over the 4-bit models.

\begin{figure*}
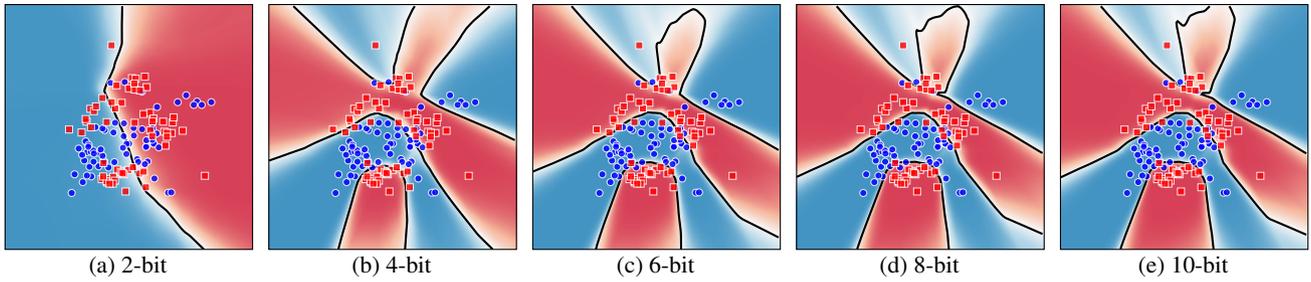

	\centering
	\def\datapath{.}
	\setlength{\figurewidth}{.19\textwidth}
	\setlength{\figureheight}{\figurewidth}
	\pgfplotsset{
	  scale only axis,
	  axis on top,
	  ytick=\empty,        %
	  yticklabels=\empty,  %
	  xtick=\empty,        %
	  xticklabels=\empty,  %
	}
	\begin{subfigure}[t]{.19\textwidth}
	  \centering
	  \input{2_bit.tex}\\[-.5em]
	  \caption{2-bit}
	\end{subfigure}
	\hfill
	\begin{subfigure}[t]{.19\textwidth}
	  \centering
	  \input{4_bit.tex}\\[-.5em]
	  \caption{4-bit}
	\end{subfigure}
	\hfill
	\begin{subfigure}[t]{.19\textwidth}
	  \centering
	  \input{6_bit.tex}\\[-.5em]
	  \caption{6-bit}
	\end{subfigure}
	\hfill
	\begin{subfigure}[t]{.19\textwidth}
	  \centering
	  \input{8_bit.tex}\\[-.5em]
	  \caption{8-bit}
	\end{subfigure}
	\hfill
	\begin{subfigure}[t]{.19\textwidth}
	  \centering
	  \input{10_bit.tex}\\[-.5em]
	  \caption{10-bit}
	\end{subfigure}
	\caption{\textbf{Chopping the banana:} We start from a 10-bit NN \ours\ results on the Banana binary classification data set and gradually decrease the fractional precision of the trained model. The low-bit models up to 4 bits capture the overall structure well. This is further confirmed by the results in \cref{tbl:two-moons-nlpd}.}
	\label{fig:choppedbanana}
  \end{figure*}

\subsection{Ablation Studies}
\label{sec:ablations}

\begin{figure}[h!]
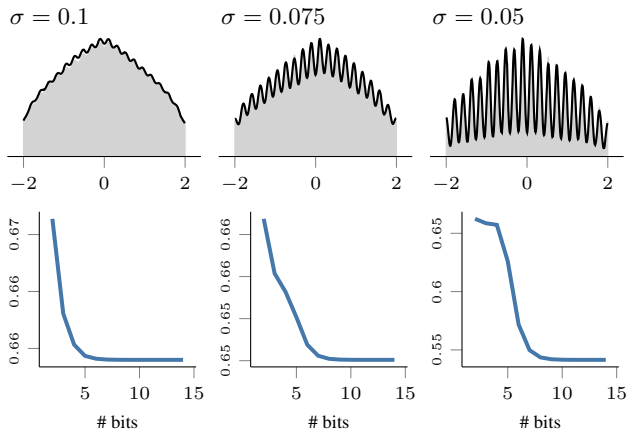

	\centering
	\setlength{\figurewidth}{0.5\columnwidth}
	\setlength{\figureheight}{0.4\columnwidth}
	\begingroup
	\pgfplotsset{
	  axis on top,
      axis y line=none,    %
      axis x line*=bottom,
      tick align=outside,
      ymin=0,
      clip=false,
      ytick=\empty,        %
      yticklabels=\empty,  %
      xticklabel style={font=\scriptsize},
    }
	\newcommand{\mylabel}[1]{%
	  \tikz[baseline]\node[anchor=base,font=\bf\footnotesize,align=left,text width=.6\figurewidth,yshift=0.5cm]{#1};\\[-1em]}
    \hfill
    \begin{minipage}[t]{.3\columnwidth}
      \raggedleft\mylabel{$\sigma = 0.1$}
      \input{smoothness-mixture-0.1sigma-density.tex} %
    \end{minipage}
    \hfill
    \begin{minipage}[t]{.3\columnwidth}
      \raggedleft\mylabel{$\sigma = 0.075$}
      \input{smoothness-mixture-0.075sigma-density.tex} %
    \end{minipage}
    \hfill
    \begin{minipage}[t]{.3\columnwidth}
	  \raggedleft\mylabel{$\sigma = 0.05$}
      \input{smoothness-mixture-0.05sigma-density.tex} %
    \end{minipage}
	\hfill\\[-1em]
	\endgroup
	\begingroup
	\setlength{\figurewidth}{0.25\columnwidth}
	\setlength{\figureheight}{0.25\columnwidth}
	\pgfplotsset{
      axis x line*=bottom,
      axis y line*=left,
      tick align=outside,
      clip=false,
      xticklabel style={font=\scriptsize},
      yticklabel style={font=\tiny,rotate=90},      
      xlabel={\scriptsize \# bits},
      scale only axis,
      semithick/.append style={line width=1.5pt},
    }
	\hfill
    \begin{minipage}[t]{.3\columnwidth}
      \raggedleft
\begin{tikzpicture}

\begin{axis}[
ymin=0.653375616669655, ymax=0.667001554369926,
width=\figurewidth,
height=\figureheight
]
\addplot [semithick, scBlue]
table {%
2 0.666382193565369
3 0.658052921295166
4 0.655324399471283
5 0.654340982437134
6 0.654094457626343
7 0.654020309448242
8 0.654001116752625
9 0.653996229171753
10 0.653995037078857
11 0.653995037078857
12 0.653994977474213
13 0.653994977474213
14 0.653995275497437
};
\end{axis}

\end{tikzpicture} %
    \end{minipage}
    \hfill
    \begin{minipage}[t]{.3\columnwidth}
      \raggedleft
\begin{tikzpicture}

\begin{axis}[
ymin=0.649260973930359, ymax=0.667703652381897,
width=\figurewidth,
height=\figureheight
]
\addplot [semithick, scBlue]
table {%
2 0.666865348815918
3 0.660378575325012
4 0.65819263458252
5 0.655171990394592
6 0.651889443397522
7 0.650584876537323
8 0.650222063064575
9 0.650128901004791
10 0.65010529756546
11 0.650099396705627
12 0.650099337100983
13 0.650099277496338
14 0.650099635124207
};
\end{axis}

\end{tikzpicture} %
	\end{minipage}
	\hfill
    \begin{minipage}[t]{.3\columnwidth}
      \raggedleft
\begin{tikzpicture}

\begin{axis}[
ymin=0.535424107313156, ymax=0.668442732095718,
width=\figurewidth,
height=\figureheight
]
\addplot [semithick, scBlue]
table {%
2 0.662396430969238
3 0.658631801605225
4 0.657230019569397
5 0.626076757907867
6 0.571706712245941
7 0.549807548522949
8 0.543579697608948
9 0.541991174221039
10 0.541590094566345
11 0.541490077972412
12 0.541470408439636
13 0.541470408439636
14 0.541470527648926
};
\end{axis}

\end{tikzpicture} %
    \end{minipage}
	\hfill
    \endgroup\\[-2em]
  \caption{Ablation result of \ours\ (black) for target distributions with increasing complexity (gray)  and the precision used by the variational distribution to represent the target. The entropy (lower figures) shows the cut-off for bitstring depth needed to represent each target.}
  \label{fig:ablation}
\end{figure}

\paragraph{Increasing Complexity of Target Distribution}
We consider an ablation study where we control the target distribution complexity. 
For this, we constructed a mixture of equidistant Gaussians and assessed the entropy of \ours\ under varying numbers of bits under three different amounts of variance for each Gaussian.
\cref{fig:ablation} shows the fitted results of \ours\ (black) with 16 bits for target distributions with increasing complexity (gray) alongside the entropy of \ours\ under varying number of bits. The entropy (lower figures) shows the cut-off for number of bits needed to represent each target, indicating that \ours\ naturally exhibits a parsimonious behaviour.

\paragraph{Trade-off Between Model Complexity and Bitstring Depth}
For NN applications, an interesting question is whether fine-grained numerical accuracy is needed to represent the model weights in the first place. Recent advances in large-scale model training and inference suggest that rather than numerical accuracy, the models benefit from more parameters, which enable further flexibility. Hence, we study whether the models benefit from higher numerical granularity w.r.t.\ probabilistic treatment. 

In \cref{tbl:two-moons-nlpd}, we vary both the NN complexity (units in the two hidden layers) and the bitstring length. We consider 2--12-bit models (with only fractional bits). The negative log predictive density (NLPD, smaller better) on the two moons data suggests that even low bit depth models perform well, and the dominating factor in expressivity is the number of units in the NN. In \cref{app:results}, we include similar tables for both accuracy and expected calibration error (ECE).

\paragraph{Do Bitstrings Capture Hierarchies in NNs?}
Finally, we use a neural network model to study the hierarchies captured by \ours. We start from a 10-bit NN \ours\ results on the Banana binary classification data set and gradually decrease the fractional precision of the trained model, chopping off more granular levels of the model. \cref{fig:choppedbanana} shows the results for 10, 8, 6, 4, and 2-bit models (2 integer bits each, except for the 2-bit model). Even the 4-bit model (2 integer bits and 1 fractional bit) captures the overall structure well, whereas the 2-bit model (with no integer bits; only a sign bit and a fraction bit) struggles. 

\begin{table}[t]
	\centering\scriptsize
	\caption{The trade-off between NN model complexity (units in hidden layers) and bitstring length (2--12 bits). The negative log predictive density (NLPD, smaller better) on the two moons data suggests that even low bit depth models perform well, and the dominating factor in expressivity is the number of units in the NN. See \cref{app:results} for ACC/ECE.}
	\label{tbl:two-moons-nlpd}
	\vspace*{-.5em}

	\setlength{\tabcolsep}{4.5pt}

	\newcommand{\scaleFactor}{0.20} %
	\newcommand{\minFactor}{0.20} %

	\newcommand{\val}[3]{%
	  \pgfmathsetmacro\perc{100-min(100, max(0, ((#2-\minFactor)/\scaleFactor)*100))} 
	  \edef\computedcolor{\noexpand\cellcolor{scCyan!\perc}}%
	  \computedcolor \pgfmathprintnumber[fixed,precision=2]{#2}%
	}  
	
	\newcommand{\rotatecol}[1]{\makebox[1em][l]{\rotatebox{90}{#1}}}
\begin{tabular}{llccccccc}
\toprule
 && \multicolumn{7}{c}{Increasing NN complexity $\rightarrow$} \\
 && [4, 4] & [6, 6] & [8, 8] & [10, 10] & [12, 12] & [14, 14] & [16, 16] \\ 
\midrule
\multirow{10}{*}{\rotatecol{Bitstring depth}}
&2 & \val{}{0.35737}{0.02589} & \val{}{0.35085}{0.02077} & \val{}{0.35420}{0.02063} & \val{}{0.31669}{0.00914} & \val{}{0.32610}{0.03645} & \val{}{0.29647}{0.01496} & \val{}{0.29481}{0.02885} \\ 
&3 & \val{}{0.36568}{0.01758} & \val{}{0.36355}{0.01681} & \val{}{0.26115}{0.02400} & \val{}{0.33643}{0.04315} & \val{}{0.27068}{0.04537} & \val{}{0.24457}{0.01588} & \val{}{0.25085}{0.00363} \\ 
&4 & \val{}{0.37655}{0.00696} & \val{}{0.31726}{0.03391} & \val{}{0.31050}{0.05267} & \val{}{0.29510}{0.06752} & \val{}{0.26790}{0.05159} & \val{}{0.28480}{0.05728} & \val{}{0.24460}{0.01634} \\ 
&5 & \val{}{0.35144}{0.03800} & \val{}{0.32130}{0.05350} & \val{}{0.36248}{0.01725} & \val{}{0.29344}{0.05994} & \val{}{0.27281}{0.05049} & \val{}{0.25040}{0.02131} & \val{}{0.24847}{0.00254} \\ 
&6 & \val{}{0.34243}{0.04214} & \val{}{0.34353}{0.05242} & \val{}{0.36883}{0.00903} & \val{}{0.30094}{0.06153} & \val{}{0.27592}{0.05655} & \val{}{0.25470}{0.01848} & \val{}{0.24335}{0.00810} \\ 
&7 & \val{}{0.31305}{0.04045} & \val{}{0.30337}{0.05961} & \val{}{0.29631}{0.06657} & \val{}{0.26335}{0.03306} & \val{}{0.27810}{0.05866} & \val{}{0.24559}{0.02082} & \val{}{0.24330}{0.01165} \\ 
&8 & \val{}{0.32853}{0.04618} & \val{}{0.31049}{0.06242} & \val{}{0.25309}{0.01973} & \val{}{0.29744}{0.06448} & \val{}{0.28749}{0.04768} & \val{}{0.25500}{0.02291} & \val{}{0.25615}{0.02582} \\ 
&9 & \val{}{0.35551}{0.03283} & \val{}{0.32488}{0.05912} & \val{}{0.31995}{0.06012} & \val{}{0.32726}{0.06195} & \val{}{0.26361}{0.04020} & \val{}{0.23338}{0.00652} & \val{}{0.24978}{0.03222} \\ 
&10 & \val{}{0.33439}{0.04965} & \val{}{0.34664}{0.05042} & \val{}{0.29637}{0.05545} & \val{}{0.30456}{0.06037} & \val{}{0.24720}{0.02448} & \val{}{0.26357}{0.05707} & \val{}{0.23597}{0.00875} \\ 
&12 & \val{}{0.37084}{0.00232} & \val{}{0.28581}{0.05037} & \val{}{0.34834}{0.05062} & \val{}{0.34967}{0.04381} & \val{}{0.26280}{0.05501} & \val{}{0.26864}{0.05733} & \val{}{0.24079}{0.00836} \\ 
\bottomrule
\end{tabular}

\end{table}

\section{Discussion and Conclusion}
\label{sec:discussion}
In this work, we introduced \textbf{\ours}, a novel approach for approximate Bayesian inference that operates directly in the space of discrete bitstring representations. By leveraging (deterministic) probabilistic circuits as the representational framework, we demonstrated that inference can be performed directly on bitstring representations of number systems, enabling effective approximate inference and uncertainty quantification.
Our approach presents a paradigm shift by learning a rich variational approximation induced by a variational family on bitstring representations without relying on high-precision representations.
Our experiments showcased the flexibility of \ours\ across different settings: In \cref{sec:2d-densities}, we illustrated its ability to approximate complex non-Gaussian densities; and in \cref{sec:mlp-exp}, we demonstrated its effectiveness in Bayesian deep learning, where it provided robust uncertainty estimates while maintaining computational efficiency. 

Beyond demonstrating feasibility, these results highlight that flexible approximate Bayesian inference does not need to be constrained to continuous-valued computations but can be reformulated in a fully discrete manner. 
Moreover, our results further highlight the potential of using probabilistic circuits as the representational framework for approximate inference. A key strength derived from using probabilistic circuits is the tractability of \ours\, which we exploited for computing the entropy of our model in closed-form.
While \ours\ provides a promising direction for flexible variational inference, several limitations remain.

\paragraph{Limitations}
To scale to high-dimensional settings, our approach employs a mean-field approximation to the posterior.
This limitation arises from our tree construction, which considers dependencies between all bits and dependencies between all dimensions if no mean-field assumption is made.
In practical applications, modeling all dependencies is likely unnecessary and introduces an excessive computational and memory burden.
Therefore, a promising future direction is to leverage more compact representations such \citep{peharz2020einsum}.
For the same reason, our approach also currently introduces many parameters to be optimized, which can result in further challenges for high-dimensional settings.
Lastly, our experiments focused only on fixed-point representations. Exploiting the representational power of floating-point representations is a promising future avenue.

The codes and resources for \ours\ are available on GitHub: \url{github.com/AaltoML/bitvi}.

\begin{acknowledgements} 
A.\ Solin acknowledges funding from the Research Council of Finland (grant number 339730). M.\ Trapp acknowledges funding from the Research Council of Finland (grant number 347279). A.\ Sladek acknowledges funding from the Finnish Doctoral Program Network in Artificial Intelligence (AI-DOC, decision number VN/3137/2024-OKM-6). We acknowledge
the computational resources provided by the Aalto Science-IT project. We thank the reviewers and the area chair for their constructive feedback.
\end{acknowledgements}

\newpage

\onecolumn

\title{Approximate Bayesian Inference via Bitstring Representations\\(Supplementary Material)}
\maketitle

\appendix

\section{Technical Details}\label{app:details}

\subsection{Probabilistic Circuits}
We will briefly review the main concepts related to probabilistic circuits (PC), relevant for this work.

\begin{definition}[scope of a node]
The scope of a node $\node$ is the set of variables it depends on. The scope for a given node $\node$ is denoted as $\scope\node$. See \citep{trapp2019bayesian} for details.
\end{definition}

\begin{definition}[support of a node]
	The support of a node $\node(\bx)$ is notated as $\supp (\node)$, and is defined as $\supp (\node) = \{\bx \in \mcX \,|\, \node(\bx) > 0\}$. It is the set of values in $\mcX$ for which the node computes a non-zero value.
\end{definition}

\begin{definition}[smooth \& decomposable circuit]
A sum node is smooth if its children have the same scope.
A product node is decomposable if its children have pairwise disjoint scopes.
A circuit is smooth (resp. decomposable) if all its sum nodes are smooth (resp. product nodes are decomposable).
\end{definition}

In this work, we only consider circuits that fullfill both smoothness and decomposability conditions as they both are required to render common inference tasks, such as density evaluation and marginalisation, tractable.

\subsection{Fixed-point Representation Bitstrings} \label{app:fp-mapping}

In this work we focus on fixed-point bitstrings. A $B$-bit bitstring $\bb = \{b_{(B-1)}, b_{(B-2)}, \ldots , b_{B_0}\}$ is defined as having $1$ sign bit, $F$ fractional bits and $B - F - 1$ integer bits. The mapping function $\phi$ is defined as
\[
	\phi(\bb) = (-1)^{b_k}\left(\sum_{i=0}^{k-1}b_i 2^{i - F}\right) .
\]

Fixed-point bitstrings encode fractional values as negative powers of two, and integer values as positive powers of two.

\subsection{Multivariate Bitstring Representations} \label{app:mv-details}
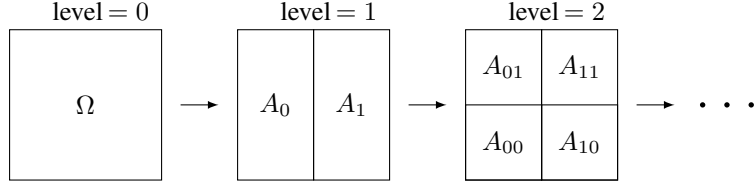
\begin{figure}
\centering
  		\begin{tikzpicture}
		\draw[draw=black] (0,0) rectangle ++(2, 2);
		\node[] at (1, 1) {$\Omega$};
    \node[] at (1.2, 2.25) {level$\,=0$};
		
		\draw[-latex]  (2.25, 1) -- (2.75, 1);
		
		\begin{scope}[xshift=3cm]
			\draw[draw=black] (0,0) rectangle ++(1, 2);
			\node at (.5, 1) {$A_0$};
			\draw[draw=black] (1,0) rectangle ++(1, 2);
			\node at (1.5, 1) {$A_1$};
      \node[] at (1.2,2.25) {level$\,=1$};
	
			\draw[-latex]  (2.25, 1) -- (2.75, 1);
		
			\begin{scope}[xshift=3cm]
				\draw[draw=black] (0,0) rectangle ++(1, 2);
				\node at (.5, .5) {$A_{00}$};
				\draw[draw=black] (0,0) rectangle ++(1, 1);
				\node at (.5, 1.5) {$A_{01}$};
				\draw[draw=black] (1,0) rectangle ++(1, 2);
				\node at (1.5, .5) {$A_{10}$};
				\draw[draw=black] (1,0) rectangle ++(1, 1);
				\node at (1.5, 1.5) {$A_{11}$};
        \node[] at (1.2,2.25) {level$\,=2$};
				
				\draw[-latex]  (2.25, 1) -- (2.75, 1);
				
                \node at (3.5, 1) {\huge$\dots$};
			\end{scope}
		\end{scope}
	\end{tikzpicture}
	\caption{Illustration of the iterative axis-aligned splitting of the domain into hyper-rectangles (sub-domains) by the circuit. }
	\label{fig:multivariate-splitting}
\end{figure}

As outlined in the main text, for multivariate distributions, we generate a circuit model that represents a distribution over hyper-rectangles.
Let $\Omega$ denote the domain of the distribution, we recursively construct a dyadic partition of the domain into measurable subsets.
This process is done by selecting a splitting dimension at each level of the tree and splitting the hyper-rectangle according to the number system representation, \ie, in the middle for fixed-point numbers.
At the next level, we select a splitting dimension our of the remaining dimension (those that have not been split yet) and split the hyper-rectange accordingly. We make sure each dimension has been split in the process, before restarting the splitting.
The construction ends if each dimension has been split $B$ many times, where $B$ is the number of bits used in the number system.
\cref{fig:multivariate-splitting} illustrates the recursive splitting of the input domain $\Omega$ into sub-domains (hyper-rectangles).

\section{Derivations}\label{app:derivations}
\subsection{BitVI Deterministic Probabilistic Circuit Construction}
In this section, we provide a comprehensive account of our deterministic probabilistic circuit $\pc$ construction that defines a probability density function over the bitstring representation of a single continuous valued random variable $X$. Without loss of generality, we assume $X \in (0,1]$. Let $\mcC(x)$ represent a probabilistic circuit such that $\mcC = (\mcG, \pmb{\theta})$ where $\mcG$ is a computational graph parameterized by parameters in $\pmb{\theta}$. Hence, $X \sim \mcC$.
\\\\
Let $\mcG$ be a binary tree structure $\mcT$. The set of non-leaf nodes of $\mcT$ is denoted as $\mcN(\mcT)$, and the set of leaf nodes is denoted as $\mcL(\mcT)$. Furthermore, let every node $\node \in \mcN(\mcT)$ be a sum node $\snode$. Note that since $\mcT$ is a binary tree structure, this means that every sum node $\snode \in \mcN(\mcT)$ has exactly two children.

\begin{definition}[Binary Partitioning Tree Structure]
 In this work we utilize a binary tree structure $\mcT$ that serves as the computational graph of our model $\pc$. A key aspect of this structure is that it represents a finite-depth partitioning of the support of the distribution that $\pc$ represents. Let us consider a case where $\pc$ has support $\supp(\pc) = A = (a, b],\text{ where } \;a,b\in\mathbb{R}, \text{ and } \;a < b$. This entails that this is the support of the root node $\node(x)$ of $\mcT$. Next, we define the supports of the child nodes $\text{ch}(\node) = \{\node_0, \node_1\}$ of $\node$ to be a \emph{mutually exclusive partitioning} of $A$. Let $c \in A$ and $a \leq c < b$. In our case, we set $c = \frac{a + b}{2}$ (mid-point), but $c$ can be any value $x \in A$. Now, the supports of the child nodes are defined as $A_0 = \supp(\node_0) = (a, c]$ and $A_1 = \supp(\node_1) = (c, b]$. Note that $A_0 \cap A_1 = \emptyset$ and $A_0 \cup A_1 = A$. We also define the sets $\mcA_0 = \{A\}$ and $\mcA_1 = \{A^0, A^1\}$. The same procedure is then recursively applied to each partition $A_{\bb} \in \mcA_1$ to create a new set of mutually exclusive partitions $\mcA_2$. This procedure continues until the depth of the tree $\mcT$. Note that this method for partitioning the support at the mid-point of each interval arises from using the fixed-point representation system.
\end{definition}

\begin{definition}[Deterministic Sum Nodes]
Let us define each sum node $\snode \in \mcN(\mcT)$:
\[
\snode_{\bb}(x) = \begin{cases} w_L^{\bb}\node_{\bb0}(x) + w_R^{\bb}\node_{\bb1}(x) & \text{if } x \in (a^{\bb}, b^{\bb}] \\
0 & \text{else} \end{cases},
\label{eq:detsnode}\]
where $(a^{\bb},b^{\bb}] = \text{supp}(\snode_{\bb})$.
Note also that $w_L^{\bb}$ and $w_R^{\bb}$ are parameterized weights associated to sum node $\snode_{\bb}$, with the properties $w_R^{\bb} + w_L^{\bb} = 1, w_L^{\bb} > 0, w_R^{\bb} > 0$ . Note that scalars $a^{\bb}$ and $b^{\bb}$ define the \emph{support} of $\supp(\snode_{\bb})$ of node $\snode_{\bb}$, meaning $\snode_{\bb}(x)$ only has non-zero probability density within this region, as shown in \cref{eq:detsnode}. Furthermore $\int_{a^{\bb}}^{b^{\bb}} \snode_{\bb}(x) = 1$ since $w_R^{\bb} + w_L^{\bb} = 1$ and $\int_{x \in \supp(\node)} \node(x) = 1 \dee x$ for all $\node \in \mcT$. In other words, the circuit is normalized and representes a valid probability distribution at the root.

\textbf{A note on bitstring notation:} Here $\bb$ denotes a vector of bits, or a \emph{bitstring}, $b_1b_2\ldots b_d$ with commas and bracket notation omitted for notational convenience and convention. This bitstring represents the path to the current node from the root, with bit value $0$ denoting a left child and bit value $1$ a right child. bitstring $\bb = \{\}$ denotes the empty bitstring (i.e. node $\snode$), and is reserved for denoting the root node of $\mcT$. $\node_{\bb i}$ denotes a node in $\mcT$, with $i=0$ for the left child of $\snode_{\bb}$ and $i=1$ denotes the right child of $\snode_{\bb}$. $\node_{\bb i}$ may be either a $\snode$ or a $\lnode$ depending of if $\node \in \mcN(\mcT)$ or $\mcL(\mcT)$. Importantly, the child nodes of $\snode_{\bb}$ split its support $a^{\bb}$ and $b^{\bb}$ into sub-intervals $A_L = (a^{\bb 0}, b^{\bb 0}]$ and $A_R = (a^{\bb 1}, b^{\bb 1}]$, where $A_L \cap A_R = \emptyset$.

\end{definition}

\begin{definition}[Uniform Leaf Nodes]

We define every leaf node $\lnode_{\bb}(x) \in \mcL(\mcT)$ as a continuous uniform distribution:
\[
\lnode_{\bb}(x) = \begin{cases} 
\frac{1}{b^{\bb} - a^{\bb}} & \text{if } x \in (a^{\bb}, b^{\bb}]\\
0 & \text{else}
\end{cases},
\]
where $(a^{\bb},b^{\bb}] = \text{supp}(\lnode_{\bb})$.

\end{definition}

\subsection{Inverse-CDF Derivation}
This section derives the inverse cumulative density function (CDF) of our circuit construction. Recall that the inverse-CDF was presented in \cref{eq:local_moves} as a pair of local moves that can be iteratively applied at each depth of the circuit, starting from the root and traversing downwards. We begin by showing an example of the PDF and CDF for a simple circuit construction. Then, we show the computation for the inverse of the CDF function, and extend this to a more general set of rules from there.

To start, we show in the following example what the PDF of a simple (shallow) circuit construction looks like. In this case, we let $\mcT$ be a depth 3 tree structure i.e. a distribution over 3-bits. The PDF function is then

\begin{example}[PDF Example]
\[
\mcC(x) & = \snode(x) \\
        & = w_L \snode_0(x) + w_R \snode_1(x) \\
        & = w_L \left( w_L^0 \snode_{00}(x) + w_R^0 \snode_{01}(x)\right) + w_R \left( w_L^1 \snode_{10}(x) + w_R^1 \snode_{11}(x)\right) \\
        & = w_L \left( w_L^0 \left[w_L^{00} \lnode_{000}(x) + w_R^{01}\lnode_{001}(x)\right] 
        + w_R^0 \left[w_L^{01}\lnode_{010}(x) + w_R^{01}\lnode_{011}(x)\right] \right) \\ 
        & \quad + w_R \left( w_L^1 \left[w_L^{10}\lnode_{100}(x) + w_R^{10}\lnode_{101}(x)\right] + w_R^1 \left[w_L^{11}\lnode_{110}(x) + w_R^{11}\lnode_{111}(x)\right]\right)
\]
\end{example}

The PDF function can then be integrated up to some variable $x$ to define the CDF of the circuit construction. Let $F_{\pc}$ represent the CDF of $\mcC$. Then,

\begin{example}[CDF Example]
\[
F_{\pc}(x) & = \int_{0}^{x} \mcC(x) \dee x \\
 & = \int_0^x \left( w_L \snode_0(x) + w_R \snode_1(x) \right) \dee x \\
 & = w_L \int_0^x \snode_0(x) \dee x + w_R \int_0^x \snode_1(x) \dee x \\
 & = \begin{cases} w_L \int_0^x \snode_0(x) & \text{if }  x < b^0 \\ w_L + w_R\int_0^x \snode_1 (x) & \text{else} \end{cases}
\]

Note that here we leverage the facts that: $\int_0^x \snode_0(x) = 1 \text{ iff } x \geq b^0, \int_0^x \snode_1(x) \dee x = 0 \text{ iff } x < a^1$, recalling that $\supp(\snode_0(x)) = (a^0, b^0]$ and $\supp(\snode_1(x)) = (a^1,b^1]$.
\\\\
Let's consider the case $x < b^0$:
\[
F_{\pc}(x) & = w_L \int_0^x \snode_0(x) \dee x & \text{Iff } (x < b^0) \\
& = w_L \left( w_L^0 \snode_{00}(x) + w_R^0 \snode_{01}(x)\right) \\
& = w_L \left(w_L^0 \int_0^x \snode_{00}(x) \dee x + w_R^0 \int_0^x \snode_{01}(x)\dee x \right) \\
& = \begin{cases} w_L w_L^0 \int_0^x \snode_{00}(x) \dee x & \text{if } x < b^{00} \\ 
    w_L \left(w_L^0 + w_R^0 \int_0^x \snode_{01}(x) \dee x \right) & \text{else} \end{cases}
\]

Now let's consider the right branch case, $x \geq b^{00}$:

\[
F_{\pc}(x) & = w_L \left(w_L^0 + w_R^0 \int_0^x \snode_{01}(x) \dee x \right) & \text{iff } (b^{00} \leq x < b^0) \\
& = w_L \left(w_L^0 + w_R^0 \int_0^x \left[ w_L^{01}\lnode_{010}(x) + w_R^{01}\lnode_{011}(x) \right] \dee x \right) \\
& = w_L \left(w_L^0 + w_R^0 \left[ w_L^{01} \int_0^x \lnode_{010}(x) \dee x + w_R^{01} \int_0^x \lnode_{011}(x) \dee x \right] \right) \\
& = \begin{cases} w_L \left(w_L^0 + w_R^0 \left[ w_L^{01} \int_0^x \lnode_{010}(x) \dee x \right] \right) & \text{if } x < b^{010} \\
w_L \left(w_L^0 + w_R^0 \left[ w_L^{01} + w_R^{01} \int_0^x \lnode_{011}(x) \dee x \right] \right) & \text{else}
\end{cases}
\]

Considering the right branch case again, $x \geq b^{010}$:
\[
F_{\pc}(x) & = w_L \left(w_L^0 + w_R^0 \left[ w_L^{01} + w_R^{01} \int_0^x \lnode_{011}(x) \dee x \right] \right) & \text{iff } b^{010} \leq x < b^0 \\
& = w_L \left(w_L^0 + w_R^0 \left[ w_L^{01} + w_R^{01} \left\{\frac{x - a^{011}}{b^{011} - a^{011}} \right\} \right] \right)
\]

\end{example}

We can now invert this function by writing the equation w.r.t. $u$ (with $u$ denoting the CDF value $F_{\pc}(x)$ for some $x$) instead of $x$.

\begin{example}[Inverse CDF Example]
\[
u & = w_L \left(w_L^0 + w_R^0 \left[ w_L^{01} + w_R^{01} \left\{\frac{x - a^{011}}{b^{011} - a^{011}} \right\} \right] \right) & \text{iff } F_{\pc}(b^{010}) \leq u < F_{\pc}(b^0) \\
\frac{u}{w_L} & = w_L^0 + w_R^0 \left( w_L^{01} + w_R^{01} \left[\frac{x - a^{011}}{b^{011} - a^{011}} \right] \right) \\
\frac{u}{w_L} - w_L^0 & = w_R^0 \left( w_L^{01} + w_R^{01} \left[\frac{x - a^{011}}{b^{011} - a^{011}} \right] \right) \\
\frac{\frac{u}{w_L} - w_L^0}{w_R^0} & = w_L^{01} + w_R^{01} \left(\frac{x - a^{011}}{b^{011} - a^{011}} \right) \\
\frac{\frac{u}{w_L} - w_L^0}{w_R^0} - w_L^{01} & = w_R^{01} \left(\frac{x - a^{011}}{b^{011} - a^{011}}\right) \\
\frac{\frac{\frac{u}{w_L} - w_L^0}{w_R^0} - w_L^{01}}{w_R^{01}} &= \frac{x - a^{011}}{b^{011} - a^{011}} \\
x &=  (b^{011} - a^{011})\frac{\frac{\frac{u}{w_L} - w_L^0}{w_R^0} - w_L^{01}}{w_R^{01}} + a^{011}
\]
Rewriting
\[
x = (b^{011} - a^{011})\left\{\frac{1}{w_R^{01}}\left[\frac{1}{w_R^{0}} \left(\frac{1}{w_L}u - w_L^0\right) - w_L^{01} \right]\right\} + a^{011}
\]

\end{example}
Observing the structure of the derivation, this can be generalized and written as the following set of recursive rules. Letting $F^{-1}_{\pc}$ denote the inverse CDF of the circuit with $F^{-1}_{\node_{\bb}}$ denoting the inverse CDF of a given node,

\begin{definition}[Recursive Definition of the Inverse CDF]
\[
F^{-1}_{\node_{\bb}}(u) = \begin{cases} 
F^{-1}_{\snode_{\bb}}(u) & \text{if} \quad \node \in \mcN(\mcT)\\
F^{-1}_{\lnode_{\bb}}(u) & \text{if} \quad \node \in \mcL(\mcT)
\end{cases}
\]

where
\[
F^{-1}_{\snode_{\bb}}(u) = \begin{cases}
F^{-1}_{\node_{\bb 0}}\left(\frac{u}{w_L^{\bb}}\right) & \text{if} \quad u < w_L^{\bb} \\
F^{-1}_{\node_{\bb 1}}\left(\frac{u - w_L^{\bb}}{w_R^{\bb}}\right) & \text{otherwise}
\end{cases}
\]

and 
\[
F^{-1}_{\lnode_{\bb}}(u) = a^{\bb} + u(b^{\bb} - a^{\bb})
\]
\\\\
Noting that this is simply the inverse-CDF of a uniform distribution. Further note how the conditions arise for the case of sum nodes. It is the case that $u < w_L^{\bb} \equiv u < F^{-1}_{\snode_{\bb}}(b^{\bb0})$. This is due to the fact that $F^{-1}_{\snode_{\bb}}(b^{\bb 0}) = w_L^{\bb}F^{-1}_{\node_{\bb 0}} + w_R^{\bb}F^{-1}_{\node_{\bb 1}}(b^{\bb 0})$, where $F^{-1}_{\node_{\bb 0}}(b^{\bb 0}) = 1$ and $F^{-1}_{\node_{\bb 1}}(b^{\bb 0}) = 0$.
\end{definition}

\subsection{Entropy Derivation}\label{sec:entropy}

The general definition for the entropy of a probability distribution $p(x)$ is
\[
\ent{p(x)} = -\int_{x \in \mcX} p(x) \log(p(x)) \dee x
\]

Applying this to the definition of our model $\pc$,
\[
\ent{\mcC(x)} & = -\int_{x \in \mcX} \mcC(x) \log (\pc (x)) \dee x \\
& = -\int_{x \in \mcX} \snode(x) \log(\snode(x)) \dee x \\
 & =  -\int_{x \in \mcX} \left( w_L \node_0(x) + w_R \node_1(x) \right) \log (w_L \node_0(x) + w_R \node_1(x)) \dee x \\
 & =  - w_L \int_{x \in \mcX} \node_0(x) \log (w_L \node_0(x) + w_R \node_1(x)) \dee x \\ 
 & \quad - w_R \int_{x \in \mcX} \node_1(x) \log (w_L \node_0(x) + w_R \node_1(x)) \dee x \nonumber
\]

Note that due to determinism of $\pc$ and subsequently every $\node_\epsilon$ in $\pc$, the terms inside the logarithm can be simplified. In this case, $\supp(\node) = \mcX$. Then by our construction, $\supp(\node_0) = A_0,\, \supp(\node_1) = A_1$ where $A_0 \bigcup A_1 = \mcX$ and $A_0 \bigcap A_1 = \emptyset$. As such, the integrals and terms inside the logarithms simplify as follows:

\[
 \ent{\pc(x)} = -w_L \int_{x \in A_0} \node_0(x) \log (w_L \node_0(x)) \dee x - w_R \int_{x \in A_1} \node_1(x) \log (w_R \node_1(x)) \dee x
\]

since $\node_0(x) = 0 \text{ for all } x \in A_1,$ and $\node_1(x)=0 \text{ for all } x \in A_0$. Simplifying further,

\[
 \ent{\pc(x)} &= -w_L \int_{x \in A_0} \node_0(x) [\log(w_L) + \log (\node_0(x))] \dee x \\ 
& \quad - w_R \int_{x \in A_1} \node_1(x) [\log(w_R) + \log ( \node_1(x))] \dee x \nonumber \\
&= -w_L \int_{x \in A_0} \node_0(x) \log(w_L) \dee x - w_L \int_{x \in A_0} \node_0(x) \log (\node_0(x)) \dee x \\ 
& \quad - w_R \int_{x \in A_1} \node_1(x) \log(w_R) \dee x - w_R \int_{x \in A_1} \node_1(x) \log ( \node_1(x)) \dee x \nonumber \\
&= -w_L \log(w_L) \int_{x \in A_0} \node_0(x) \dee x - w_L \int_{x \in A_0} \node_0(x) \log (\node_0(x)) \dee x \\ 
& \quad - w_R  \log(w_R) \int_{x \in A_1} \node_1(x) \dee x - w_R \int_{x \in A_1} \node_1(x) \log ( \node_1(x)) \dee x \nonumber
\]

Since our circuit construction represents a normalized distribution, this means that for any $\node \in \mcT$, $\int_{x \in \supp(\node)} \node(x) = 1$. This leads to the further simplification of integral terms,
\[
\ent{\pc(x)} &= -w_L \log(w_L) - w_L \int_{x \in A_0} \node_0(x) \log (\node_0(x)) \dee x \\ 
& \quad - w_R  \log(w_R) - w_R \int_{x \in A_1} \node_1(x) \log ( \node_1(x)) \dee x . \nonumber
\]

Looking at the remaining integral terms, observe that these correspond to the entropy formulas of the child nodes. Hence, the above can be rewritten as
\[
\ent{\pc(x)} &= - w_L \log(w_L) + w_L \ent{\node_0(x)} - w_R  \log(w_R) + w_R \ent{\node_1(x)} . \nonumber
\]

If $\node_0$ and $\node_1$ are sum nodes, then $\ent{\node_0(x)}$ and $\ent{\node_0(x)}$ can be derived using the same approach. If $\node_0$ and $\node_1$ are leaf nodes, then the entropy corresponds to the closed form solution for the entropy of a uniform distribution. Put into a recursive definition:

\begin{definition}[Recursive Definition of the Entropy]
The entropy of a deterministic probabilistic circuit $\pc$ is defined as
\[ 
	\ent{\pc(x)} = \begin{cases}
		\ent{\snode_{\bb}(x)} & \text{if } \node \in \mcN(\mcT) \\
		\ent{\lnode_{\bb}(x)} & \text{if } \node \in \mcL(\mcT)
	\end{cases}
\]
where $\node$ denotes the root node of $\pc$. Furthermore,
\[
	\ent{\snode_{\bb}(x)} &=  w_L^{\bb} \log(w_L^{\bb}) + w_L^{\bb} \ent{\node_{\bb0}(x)} \\ & \quad - w_R^{\bb}  \log(w_R^{\bb}) + w_R^{\bb} \ent{\node_{\bb1}(x)} \nonumber
\]
and 
\[
	\ent{\lnode_{\bb}(x)} = \log (b^{\bb} - a^{\bb})
\]

where $\node_{\bb0}$ and $\node_{\bb1}$ are the child nodes of $\node_{\bb}$ (in the case that $\node_{\bb}$ is a sum node $\snode$), and $a^{\bb}, b^{\bb}$ are the endpoints of the interval over which $\node_{\bb}$ is defined (in the case that $\node_{\bb}$ is a leaf node $\lnode_{\bb}(x) = \frac{1}{b^{\bb} - a^{\bb}}$).

\end{definition}

\subsection{Reverse KL Divergence Calculation}
\label{sec:rkl}
Let us define a density $q$ and a density $p$. The reverse KL divergence of $q$ from $p$ is denoted as $\KL(q \,\|\, p)$, and defined as:
\[
\KL(q \,\|\, p) &= \int q(\bx) \log \frac{p(\bx)}{q(\bx)} \dee \bx \\
&= -\int q(\bx) \log q(\bx) \dee \bx + \int q(\bx) \log p(\bx) \,\dee \bx. 
\]
Note that $-\int q(\bx) \log q(\bx) \,\dee \bx$ is the entropy of distribution $q$, and will be denoted as $-\ent{q}$:
\[
\KL(q \,\|\, p) &= - \int q(\bx) \log p(\bx) \,\dee \bx -\ent{q}.
\]
Note also that $\int q(\bx) \log p(\bx) \,\dee$ is the expected value of the log-likelihood of $p$ w.r.t.\ $q$:
\[
   \KL(q \,\|\, p) &= -\EE_{\bx \sim q}\left[ \log p(\bx) \right] - \ent{q} .
\]

\section{Experimental Details} \label{app:experiments}

\subsection{2D Densities}

We present results for 2D non-Gaussian target distributions. In \cref{fig:densities-app}, we include additional results for typical benchmark target densities (mixture, Neal's funnel, two-modal Gaussian, ring, and banana) that we approximate with 4-bit/8-bit \ours, which captures the overall density and cross-dependencies well.

\subsection{MLP Neural Network Models}
\label{app:exp-MLP}
The experiments with the \emph{Bayesian-benchmarks} data sets used the following hyperparameters and setup:
\begin{itemize}
	\item NN hyperparameters
	\begin{itemize}
		\item Hidden layer size $16{\times}16$ for $D \leq 500$ and $32{\times}32$ for $D > 500$
		\item LayerNorm \citep{ba2016layernorm} applied to hidden layers (pre-activation)
	\end{itemize}
	\item PC hyperparameters
	\begin{itemize}
			\item Weight representations used two integer bits, except for the 2-bit model, which used zero integer bits
			\item Depth-based regularization for circuit parameters $\epsilon\,d^2$ with $\epsilon=0.1$
			\item Circuit weights were initialized from a beta distribution based on the height of the sum node in the circuit. The beta distribution $\alpha$ and $\beta$ were set as $2^h$ where $h$ is the height of the sum node in the circuit.
	\end{itemize}
	\item Training hyperparameters
	\begin{itemize}
		\item Adam optimizer with a learning rate of 0.001
		\item Batch size of 32 for $D \leq 500$ and 128 for $D > 500$
		\item 64 samples for computing the Monte Carlo approximation of the posterior log-joint
		\item Early stopping based on the validation set ELBO loss after 2000 epochs
		\item 5-fold cross-validation into train and test sets
		\item Validation set split from the train set with 20\% of the train set data
	\end{itemize}
\end{itemize}

\subsection{Ablation Studies}

\paragraph{Banana Chopping}

\begin{itemize}
	\item NN hyperparameters
	\begin{itemize}
		\item LayerNorm \citep{ba2016layernorm} applied to hidden layers (pre-activation)
	\end{itemize}
	\item PC hyperparameters
	\begin{itemize}
		\item Weight representations used 10 bits with no integer bits. A sign bit and nine fractional bits.
		\item Depth-based regularization for circuit parameters $\epsilon\,d^2$ with $\epsilon=0.001$.
		\item Circuit weights were initialized from a beta distribution based on the height of the sum node in the circuit. The beta distribution $\alpha$ and $\beta$ were set as $2^h$ where $h$ is the height of the sum node in the circuit.
	\end{itemize}
	\item Training hyperparameters
	\begin{itemize}
		\item Training set of 2048 points
		\item Validation set of 512 points
		\item Adam optimizer with a learning rate of 0.01
		\item Batch size of 256
	\end{itemize}

\end{itemize}

\section{Additional Results} \label{app:results}
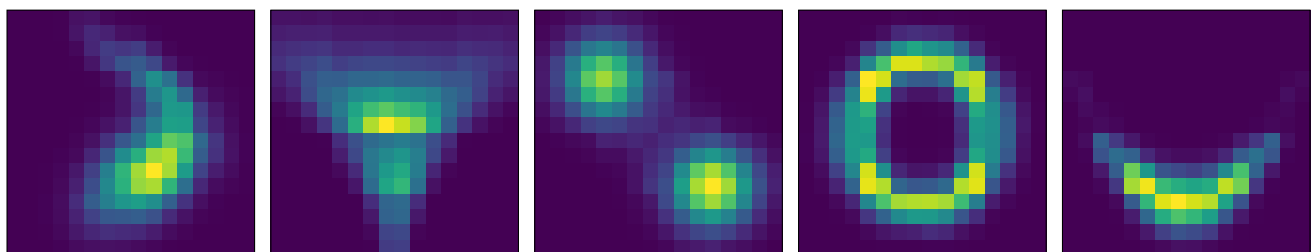
\begin{figure*}
	\centering
	\def\datapath{.}
	\setlength{\figurewidth}{.19\textwidth}
	\setlength{\figureheight}{\figurewidth}
	\pgfplotsset{
	  scale only axis,
	  axis on top,
	  ytick=\empty,        %
	  yticklabels=\empty,  %
	  xtick=\empty,        %
	  xticklabels=\empty,  %
	}
	
	{\bf Target toy 2D density functions}\\[.5em]	
	\begin{subfigure}[t]{.19\textwidth}
	  \centering
\begin{tikzpicture}

\begin{axis}[
height=\figureheight,
tick pos=left,
width=\figurewidth,
xmin=-4, xmax=3.90000000000001,
ymin=-4, ymax=3.90000000000001
]
\addplot graphics [includegraphics cmd=\pgfimage,xmin=-4, xmax=3.90000000000001, ymin=-4, ymax=3.90000000000001] {true_mixture-000.png};
\end{axis}

\end{tikzpicture}
	\end{subfigure}
	\hfill
	\begin{subfigure}[t]{.19\textwidth}
	  \centering
\begin{tikzpicture}

\begin{axis}[
height=\figureheight,
tick pos=left,
width=\figurewidth,
xmin=-4, xmax=3.90000000000001,
ymin=-4, ymax=3.90000000000001
]
\addplot graphics [includegraphics cmd=\pgfimage,xmin=-4, xmax=3.90000000000001, ymin=-4, ymax=3.90000000000001] {true_neals_funnel-000.png};
\end{axis}

\end{tikzpicture}
	\end{subfigure}
	\hfill
	\begin{subfigure}[t]{.19\textwidth}
	  \centering
\begin{tikzpicture}

\begin{axis}[
height=\figureheight,
tick pos=left,
width=\figurewidth,
xmin=-4, xmax=3.90000000000001,
ymin=-4, ymax=3.90000000000001
]
\addplot graphics [includegraphics cmd=\pgfimage,xmin=-4, xmax=3.90000000000001, ymin=-4, ymax=3.90000000000001] {true_double_gauss-000.png};
\end{axis}

\end{tikzpicture}
	\end{subfigure}
	\hfill
	\begin{subfigure}[t]{.19\textwidth}
	  \centering
\begin{tikzpicture}

\begin{axis}[
height=\figureheight,
tick pos=left,
width=\figurewidth,
xmin=-4, xmax=3.90000000000001,
ymin=-4, ymax=3.90000000000001
]
\addplot graphics [includegraphics cmd=\pgfimage,xmin=-4, xmax=3.90000000000001, ymin=-4, ymax=3.90000000000001] {true_ring-000.png};
\end{axis}

\end{tikzpicture}
	\end{subfigure}
	\hfill
	\begin{subfigure}[t]{.19\textwidth}
	  \centering
\begin{tikzpicture}

\begin{axis}[
height=\figureheight,
tick pos=left,
width=\figurewidth,
xmin=-4, xmax=3.90000000000001,
ymin=-4, ymax=3.90000000000001
]
\addplot graphics [includegraphics cmd=\pgfimage,xmin=-4, xmax=3.90000000000001, ymin=-4, ymax=3.90000000000001] {true_banana-000.png};
\end{axis}

\end{tikzpicture}
	\end{subfigure}\\[-.5em]
	
	{\bf Full-Covariance Gaussian VI} \\[.5em]	
	\begin{subfigure}[t]{.19\textwidth}
	  \centering
\begin{tikzpicture}

\begin{axis}[
height=\figureheight,
tick pos=left,
width=\figurewidth,
xmin=-4, xmax=3.90000000000001,
ymin=-4, ymax=3.90000000000001
]
\addplot graphics [includegraphics cmd=\pgfimage,xmin=-4, xmax=3.90000000000001, ymin=-4, ymax=3.90000000000001] {fcgauss_mixture-000.png};
\end{axis}

\end{tikzpicture}
	\end{subfigure}
	\hfill
	\begin{subfigure}[t]{.19\textwidth}
	  \centering
\begin{tikzpicture}

\begin{axis}[
height=\figureheight,
tick pos=left,
width=\figurewidth,
xmin=-4, xmax=3.90000000000001,
ymin=-4, ymax=3.90000000000001
]
\addplot graphics [includegraphics cmd=\pgfimage,xmin=-4, xmax=3.90000000000001, ymin=-4, ymax=3.90000000000001] {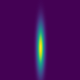};
\end{axis}

\end{tikzpicture}
	\end{subfigure}
	\hfill
	\begin{subfigure}[t]{.19\textwidth}
	  \centering
\begin{tikzpicture}

\begin{axis}[
height=\figureheight,
tick pos=left,
width=\figurewidth,
xmin=-4, xmax=3.90000000000001,
ymin=-4, ymax=3.90000000000001
]
\addplot graphics [includegraphics cmd=\pgfimage,xmin=-4, xmax=3.90000000000001, ymin=-4, ymax=3.90000000000001] {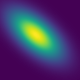};
\end{axis}

\end{tikzpicture}
	\end{subfigure}
	\hfill
	\begin{subfigure}[t]{.19\textwidth}
	  \centering
\begin{tikzpicture}

\begin{axis}[
height=\figureheight,
tick pos=left,
width=\figurewidth,
xmin=-4, xmax=3.90000000000001,
ymin=-4, ymax=3.90000000000001
]
\addplot graphics [includegraphics cmd=\pgfimage,xmin=-4, xmax=3.90000000000001, ymin=-4, ymax=3.90000000000001] {fcgauss_ring-000.png};
\end{axis}

\end{tikzpicture}
	\end{subfigure}
	\hfill
	\begin{subfigure}[t]{.19\textwidth}
	  \centering
\begin{tikzpicture}

\begin{axis}[
height=\figureheight,
tick pos=left,
width=\figurewidth,
xmin=-4, xmax=3.90000000000001,
ymin=-4, ymax=3.90000000000001
]
\addplot graphics [includegraphics cmd=\pgfimage,xmin=-4, xmax=3.90000000000001, ymin=-4, ymax=3.90000000000001] {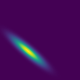};
\end{axis}

\end{tikzpicture}
	\end{subfigure}\\[-.5em]
		
    {\bf \ours\ (8-bit) result}\\[.5em]	%
	\begin{subfigure}[t]{.19\textwidth}
	  \centering
\begin{tikzpicture}

\begin{axis}[
height=\figureheight,
tick pos=left,
width=\figurewidth,
xmin=-4, xmax=3.90000000000001,
ymin=-4, ymax=3.90000000000001
]
\addplot graphics [includegraphics cmd=\pgfimage,xmin=-4, xmax=3.90000000000001, ymin=-4, ymax=3.90000000000001] {bitvi_mixture-000.png};
\end{axis}

\end{tikzpicture}\\[-.5em]
	\end{subfigure}
	\hfill
	\begin{subfigure}[t]{.19\textwidth}
	  \centering
\begin{tikzpicture}

\begin{axis}[
height=\figureheight,
tick pos=left,
width=\figurewidth,
xmin=-4, xmax=3.90000000000001,
ymin=-4, ymax=3.90000000000001
]
\addplot graphics [includegraphics cmd=\pgfimage,xmin=-4, xmax=3.90000000000001, ymin=-4, ymax=3.90000000000001] {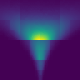};
\end{axis}

\end{tikzpicture}\\[-.5em]
	\end{subfigure}
	\hfill
	\begin{subfigure}[t]{.19\textwidth}
	  \centering
\begin{tikzpicture}

\begin{axis}[
height=\figureheight,
tick pos=left,
width=\figurewidth,
xmin=-4, xmax=3.90000000000001,
ymin=-4, ymax=3.90000000000001
]
\addplot graphics [includegraphics cmd=\pgfimage,xmin=-4, xmax=3.90000000000001, ymin=-4, ymax=3.90000000000001] {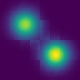};
\end{axis}

\end{tikzpicture}\\[-.5em]
	\end{subfigure}
	\hfill
	\begin{subfigure}[t]{.19\textwidth}
	  \centering
\begin{tikzpicture}

\begin{axis}[
height=\figureheight,
tick pos=left,
width=\figurewidth,
xmin=-4, xmax=3.90000000000001,
ymin=-4, ymax=3.90000000000001
]
\addplot graphics [includegraphics cmd=\pgfimage,xmin=-4, xmax=3.90000000000001, ymin=-4, ymax=3.90000000000001] {bitvi_ring-000.png};
\end{axis}

\end{tikzpicture}\\[-.5em]
	\end{subfigure}
	\hfill
	\begin{subfigure}[t]{.19\textwidth}
	  \centering
\begin{tikzpicture}

\begin{axis}[
height=\figureheight,
tick pos=left,
width=\figurewidth,
xmin=-4, xmax=3.90000000000001,
ymin=-4, ymax=3.90000000000001
]
\addplot graphics [includegraphics cmd=\pgfimage,xmin=-4, xmax=3.90000000000001, ymin=-4, ymax=3.90000000000001] {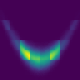};
\end{axis}

\end{tikzpicture}\\[-.5em]
	\end{subfigure}\\[.5em]
	
	{\bf \ours\ (4-bit) result}\\[.5em]	%
	\begin{subfigure}[t]{.19\textwidth}
	  \centering
\begin{tikzpicture}

\begin{axis}[
height=\figureheight,
tick pos=left,
width=\figurewidth,
xmin=-4, xmax=3.90000000000001,
ymin=-4, ymax=3.90000000000001
]
\addplot graphics [includegraphics cmd=\pgfimage,xmin=-4, xmax=3.90000000000001, ymin=-4, ymax=3.90000000000001] {bitvi_mixture_4bit-000.png};
\end{axis}

\end{tikzpicture}\\[-.5em]
	  \caption{Gaussian Mixture}
	\end{subfigure}
	\hfill
	\begin{subfigure}[t]{.19\textwidth}
	  \centering
\begin{tikzpicture}

\begin{axis}[
height=\figureheight,
tick pos=left,
width=\figurewidth,
xmin=-4, xmax=3.90000000000001,
ymin=-4, ymax=3.90000000000001
]
\addplot graphics [includegraphics cmd=\pgfimage,xmin=-4, xmax=3.90000000000001, ymin=-4, ymax=3.90000000000001] {bitvi_neals_funnel_4bit-000.png};
\end{axis}

\end{tikzpicture}\\[-.5em]
	  \caption{Neal's Funnel}
	\end{subfigure}
	\hfill
	\begin{subfigure}[t]{.19\textwidth}
	  \centering
\begin{tikzpicture}

\begin{axis}[
height=\figureheight,
tick pos=left,
width=\figurewidth,
xmin=-4, xmax=3.90000000000001,
ymin=-4, ymax=3.90000000000001
]
\addplot graphics [includegraphics cmd=\pgfimage,xmin=-4, xmax=3.90000000000001, ymin=-4, ymax=3.90000000000001] {bitvi_double_gauss_4bit-000.png};
\end{axis}

\end{tikzpicture}\\[-.5em]
	  \caption{Two-modal Gaussian}
	\end{subfigure}
	\hfill
	\begin{subfigure}[t]{.19\textwidth}
	  \centering
\begin{tikzpicture}

\begin{axis}[
height=\figureheight,
tick pos=left,
width=\figurewidth,
xmin=-4, xmax=3.90000000000001,
ymin=-4, ymax=3.90000000000001
]
\addplot graphics [includegraphics cmd=\pgfimage,xmin=-4, xmax=3.90000000000001, ymin=-4, ymax=3.90000000000001] {bitvi_ring_4bit-000.png};
\end{axis}

\end{tikzpicture}\\[-.5em]
	  \caption{Ring}
	\end{subfigure}
	\hfill
	\begin{subfigure}[t]{.19\textwidth}
	  \centering
\begin{tikzpicture}

\begin{axis}[
height=\figureheight,
tick pos=left,
width=\figurewidth,
xmin=-4, xmax=3.90000000000001,
ymin=-4, ymax=3.90000000000001
]
\addplot graphics [includegraphics cmd=\pgfimage,xmin=-4, xmax=3.90000000000001, ymin=-4, ymax=3.90000000000001] {bitvi_banana_4bit-000.png};
\end{axis}

\end{tikzpicture}\\[-.5em]
	  \caption{Banana}
	\end{subfigure}\\[-.25em]
	
	\caption{2D non-Gaussian target distributions. We include results for typical benchmark target densities (mixture, Neal's funnel, two-modal Gaussian, ring, and banana) that we approximate with 4-bit/8-bit \ours, which captures the overall density and cross-dependencies well.}
	\label{fig:densities-app}
\end{figure*}

The following section contains additional results. In \cref{fig:densities-app}, we show results of applying \ours\ in the task of approximating several benchmark non-Gaussian densities and compare the result to a baseline of a full-covariance Gaussian as the variational distribution. Furthermore, \cref{tbl:acc} and \cref{tbl:ece} provide further results (the accuracy and ECE metrics respectively) from the experiment investigating the relationship between NN model complexity and bitstring length.

\subsection{Ablation Studies}

\begin{table*}[t]
  \centering\scriptsize
  \caption{Trade-off between NN model complexity (units in hidden layers) and bitstring depth (2--12 bits). Accuracy and expected calibration error (ECE) on the two moons data suggest that even low bit depth models perform well, and the dominating factor in expressivity is the number of units in the NN. See \cref{tbl:two-moons-nlpd} in the main paper for the NLPD.}
  
  \setlength{\tabcolsep}{8pt}
  
  \begin{subfigure}{\columnwidth}
	\centering\scriptsize  
    \caption{Accuracy}
    \label{tbl:acc}

    \newcommand{\scaleFactor}{0.10} %
    \newcommand{\minFactor}{0.85} %

    \newcommand{\val}[3]{%
      \pgfmathsetmacro\perc{min(100, max(0, ((#2-\minFactor)/\scaleFactor)*100))} 
      \edef\computedcolor{\noexpand\cellcolor{scCyan!\perc}}%
      \computedcolor \pgfmathprintnumber[fixed,precision=3]{#2}%
    }  
  
    \begin{tabular}{lccccccc}
\hline
 & [4, 4] & [6, 6] & [8, 8] & [10, 10] & [12, 12] & [14, 14] & [16, 16] \\ 
\hline
2 & \val{}{0.85625}{0.01245} & \val{}{0.85000}{0.01121} & \val{}{0.85234}{0.00156} & \val{}{0.86953}{0.02218} & \val{}{0.86797}{0.01736} & \val{}{0.88828}{0.01007} & \val{}{0.88984}{0.01511} \\ 
3 & \val{}{0.85391}{0.00191} & \val{}{0.85703}{0.00911} & \val{}{0.90312}{0.00757} & \val{}{0.86484}{0.01740} & \val{}{0.89687}{0.02299} & \val{}{0.90938}{0.00870} & \val{}{0.90625}{0.00699} \\ 
4 & \val{}{0.85156}{0.00000} & \val{}{0.87891}{0.01782} & \val{}{0.88047}{0.02478} & \val{}{0.88438}{0.02724} & \val{}{0.90365}{0.02404} & \val{}{0.89766}{0.02426} & \val{}{0.90859}{0.00398} \\ 
5 & \val{}{0.86016}{0.01952} & \val{}{0.87734}{0.02478} & \val{}{0.85391}{0.00313} & \val{}{0.89453}{0.02806} & \val{}{0.90078}{0.02502} & \val{}{0.90859}{0.00398} & \val{}{0.91328}{0.00518} \\ 
6 & \val{}{0.86328}{0.02544} & \val{}{0.86068}{0.01869} & \val{}{0.85313}{0.00313} & \val{}{0.88672}{0.02902} & \val{}{0.89531}{0.02229} & \val{}{0.91016}{0.00699} & \val{}{0.90625}{0.00349} \\ 
7 & \val{}{0.88203}{0.02536} & \val{}{0.88346}{0.02566} & \val{}{0.88828}{0.03001} & \val{}{0.89922}{0.02243} & \val{}{0.89609}{0.02325} & \val{}{0.90859}{0.00765} & \val{}{0.90547}{0.00292} \\ 
8 & \val{}{0.87370}{0.02404} & \val{}{0.87695}{0.02666} & \val{}{0.90859}{0.00804} & \val{}{0.88438}{0.02724} & \val{}{0.88594}{0.02512} & \val{}{0.90938}{0.00383} & \val{}{0.90391}{0.01094} \\ 
9 & \val{}{0.86406}{0.02500} & \val{}{0.87305}{0.02904} & \val{}{0.87266}{0.02747} & \val{}{0.87656}{0.03150} & \val{}{0.89687}{0.02286} & \val{}{0.91406}{0.00552} & \val{}{0.90859}{0.00804} \\ 
10 & \val{}{0.86953}{0.02586} & \val{}{0.86198}{0.02159} & \val{}{0.88594}{0.02584} & \val{}{0.88359}{0.02688} & \val{}{0.91172}{0.00313} & \val{}{0.89922}{0.02413} & \val{}{0.90859}{0.00530} \\ 
12 & \val{}{0.85156}{0.00000} & \val{}{0.88802}{0.02646} & \val{}{0.86328}{0.01976} & \val{}{0.86250}{0.01998} & \val{}{0.89766}{0.02310} & \val{}{0.89531}{0.02201} & \val{}{0.90781}{0.00398} \\ 
\hline
\end{tabular}

  \end{subfigure}\\[2em]

  \begin{subfigure}{\columnwidth}
	\centering\scriptsize
	\caption{ECE}
	\label{tbl:ece}

	\newcommand{\scaleFactor}{0.03} %
	\newcommand{\minFactor}{0.04} %

	\newcommand{\val}[3]{%
	  \pgfmathsetmacro\perc{100-min(100, max(0, ((#2-\minFactor)/\scaleFactor)*100))} 
	  \edef\computedcolor{\noexpand\cellcolor{scCyan!\perc}}%
	  \computedcolor \pgfmathprintnumber[fixed,precision=3]{#2}%
	}  
	
	\begin{tabular}{lccccccc}
\hline
 & [4, 4] & [6, 6] & [8, 8] & [10, 10] & [12, 12] & [14, 14] & [16, 16] \\ 
\hline
2 & \val{}{0.05934}{0.01827} & \val{}{0.05318}{0.01392} & \val{}{0.06109}{0.00442} & \val{}{0.06396}{0.01429} & \val{}{0.06476}{0.01543} & \val{}{0.05499}{0.01478} & \val{}{0.05955}{0.00397} \\ 
3 & \val{}{0.06002}{0.00657} & \val{}{0.07087}{0.00746} & \val{}{0.05145}{0.01144} & \val{}{0.05309}{0.01189} & \val{}{0.04595}{0.00699} & \val{}{0.04161}{0.00344} & \val{}{0.04901}{0.00676} \\ 
4 & \val{}{0.06426}{0.00657} & \val{}{0.06432}{0.00935} & \val{}{0.05461}{0.01093} & \val{}{0.04463}{0.00882} & \val{}{0.04200}{0.00643} & \val{}{0.04547}{0.00904} & \val{}{0.03846}{0.00369} \\ 
5 & \val{}{0.06091}{0.01414} & \val{}{0.05671}{0.01500} & \val{}{0.05814}{0.00442} & \val{}{0.05255}{0.02380} & \val{}{0.04204}{0.00334} & \val{}{0.04301}{0.00948} & \val{}{0.04435}{0.00661} \\ 
6 & \val{}{0.05979}{0.01040} & \val{}{0.06666}{0.01413} & \val{}{0.05669}{0.00738} & \val{}{0.04845}{0.00759} & \val{}{0.04425}{0.00515} & \val{}{0.03856}{0.00441} & \val{}{0.04629}{0.00652} \\ 
7 & \val{}{0.06287}{0.01171} & \val{}{0.05273}{0.01647} & \val{}{0.04807}{0.00678} & \val{}{0.04631}{0.01287} & \val{}{0.04107}{0.00843} & \val{}{0.04174}{0.00347} & \val{}{0.04592}{0.00693} \\ 
8 & \val{}{0.06084}{0.01005} & \val{}{0.05088}{0.00717} & \val{}{0.03826}{0.00661} & \val{}{0.04673}{0.00629} & \val{}{0.04838}{0.01109} & \val{}{0.04517}{0.00484} & \val{}{0.04157}{0.00584} \\ 
9 & \val{}{0.05457}{0.00541} & \val{}{0.05584}{0.01120} & \val{}{0.05311}{0.00320} & \val{}{0.05312}{0.00667} & \val{}{0.04486}{0.01355} & \val{}{0.03985}{0.00470} & \val{}{0.04241}{0.00726} \\ 
10 & \val{}{0.05694}{0.01053} & \val{}{0.05773}{0.00942} & \val{}{0.05560}{0.01128} & \val{}{0.04982}{0.00219} & \val{}{0.03686}{0.00491} & \val{}{0.04651}{0.00762} & \val{}{0.04568}{0.00624} \\ 
12 & \val{}{0.06422}{0.00565} & \val{}{0.05017}{0.00914} & \val{}{0.05493}{0.01019} & \val{}{0.05255}{0.00536} & \val{}{0.04068}{0.00692} & \val{}{0.04988}{0.00955} & \val{}{0.04460}{0.00720} \\ 
\hline
\end{tabular}

  \end{subfigure}	
\end{table*}

\end{document}